%% file: main.tex
\documentclass{article} % For LaTeX2e
\usepackage{iclr2025_conference,times}
\usepackage[utf8]{inputenc} % allow utf-8 input
\usepackage[T1]{fontenc}    % use 8-bit T1 fonts
\usepackage{hyperref}       % hyperlinks
\usepackage{url}            % simple URL typesetting
\usepackage{booktabs}       % professional-quality tables
\usepackage{amsfonts}       % blackboard math symbols
\usepackage{nicefrac}       % compact symbols for 1/2, etc.
\usepackage{microtype}      % microtypography
\usepackage{xcolor}         % colors
\usepackage{graphicx}       % figures
\usepackage{subcaption}     % subfigures
\usepackage{textcomp}
\usepackage{amsmath}
\usepackage{amssymb}
\usepackage{multirow}
\usepackage{siunitx}
\usepackage{svg}
\usepackage{soul}
\usepackage{tikz}
\usetikzlibrary{shapes.misc}
\tikzset{cross/.style={cross out, draw=black, minimum size=2*(#1-\pgflinewidth), inner sep=0pt, outer sep=0pt},
cross/.default={4.5pt}}
\usepackage{enumitem}
\usepackage{algorithm}
\usepackage[noend]{algpseudocode}

\usepackage[colorinlistoftodos]{todonotes}
\usetikzlibrary{shapes,arrows,positioning,fit,calc,backgrounds,shapes.geometric}

\definecolor{babyblue}{rgb}{0.06,0.58,0.97}
\definecolor{salmon}{rgb}{0.98,0.41,0.38}

\graphicspath{ {./figures/} }

\input{math_commands.tex}

\newcommand{\figref}[1]{Figure~\ref{fig:#1}}

\newcommand{\figrefs}[2]{Figures~\ref{fig:#1}--\ref{fig:#2}}
\newcommand{\figlabel}[1]{\label{fig:#1}}

\newcommand{\sectionref}[1]{Section~\ref{section:#1}}
\newcommand{\sectionlabel}[1]{\label{section:#1}}
\newcommand{\appendixref}[1]{Appendix~\ref{appendix:#1}}
\newcommand{\appendixlabel}[1]{\label{appendix:#1}}
\newcommand{\equationref}[1]{Equation~\ref{equation:#1}}
\newcommand{\equationlabel}[1]{\label{equation:#1}}

\newcommand{\ourpipeline}{SFvODE}
\newcommand{\ourpipelinefull}{Scene Flow via ODE}

\newcommand{\ourmethod}{EulerFlow}

\newcommand{\pointcloud}{P}
\newcommand{\pointcloudt}{P_t}
\newcommand{\pointcloudsub}[1]{P_{#1}}
\newcommand{\pointcloudtpone}{P_{t+1}}
\newcommand{\pointcloudtmone}{P_{t-1}}

\newcommand{\flow}{\hat{\mathcal{F}}}
\newcommand{\flowgt}{\mathcal{F}}
\newcommand{\flowttpone}{\flow_{t,t+1}}

\newcommand{\flowgtttpone}{\flowgt_{t,t+1}}

\newcommand{\norm}[1]{\left\lVert #1 \right\rVert}

\newcommand{\poorparagraph}[1]{\textbf{{#1}.}}

\newcommand{\chamferdistancenameraw}{TruncatedChamfer}
\newcommand{\chamferdistancename}{\textup{\chamferdistancenameraw{}}}
\newcommand{\chamferdistance}[2]{\chamferdistancename(#1, #2)}

\newcommand{\network}{\theta}

\newcommand{\dirforward}{\texttt{FWD}}
\newcommand{\dirbackward}{\texttt{BWD}}

\newcommand{\kstepname}{\textup{Euler}}
\newcommand{\kstep}[2]{\kstepname_{\network}\left(#1, #2\right)}

\newcommand{\bevmaintextfontsize}{\fontsize{5pt}{5pt}\selectfont}
\DeclareCaptionFont{bevmaintextfont}{\bevmaintextfontsize}

\newcommand{\resulttablefontsize}{\fontsize{6.3pt}{6.3pt}\selectfont}
\DeclareCaptionFont{resulttablefont}{\resulttablefontsize}

\makeatletter
\DeclareRobustCommand{\iscircle}{\mathord{\mathpalette\is@circle\relax}}
\newcommand\is@circle[2]{%
  \begingroup
  \sbox\z@{\raisebox{\depth}{$\m@th#1\bigcirc$}}%
  \sbox\tw@{$#1\square$}%
  \resizebox{!}{\ht\tw@}{\usebox{\z@}}%
  \endgroup
}
\makeatother

\newcommand{\websitelink}{\href{https://vedder.io/eulerflow}{\texttt{vedder.io/eulerflow}}}

\usepackage{hyperref}
\usepackage{url}

\title{Neural Eulerian Scene Flow Fields}

\author{%
Kyle Vedder$^{1,2}$\thanks{Corresponding email: \texttt{kvedder@seas.upenn.edu}} \quad 
Neehar Peri$^{2,3}$ \quad Ishan Khatri$^{3}$ \quad Siyi Li$^{1}$ \quad Eric Eaton$^{1}$ \\ 
\textbf{Mehmet Kocamaz}$^{2}$ \quad \textbf{Yue Wang}$^{2}$ \quad \textbf{Zhiding Yu}$^{2}$ \quad \textbf{Deva Ramanan}$^{3}$ \quad \textbf{Joachim Pehserl}$^{2}$ \\
{\small$^1$University of Pennsylvania \quad $^2$NVIDIA \quad $^3$Carnegie Mellon University}
}

\iclrfinalcopy % Uncomment for camera-ready version, but NOT for submission.
\begin{document}

\maketitle

\newcommand{\captionfontsize}{\fontsize{7}{7}\selectfont}
% 2.67 x = 5.5in
\begin{figure}[h!]
    \centering
    \begin{subfigure}[b]{0.359\textwidth}
        \centering
        \includegraphics[width=\textwidth]{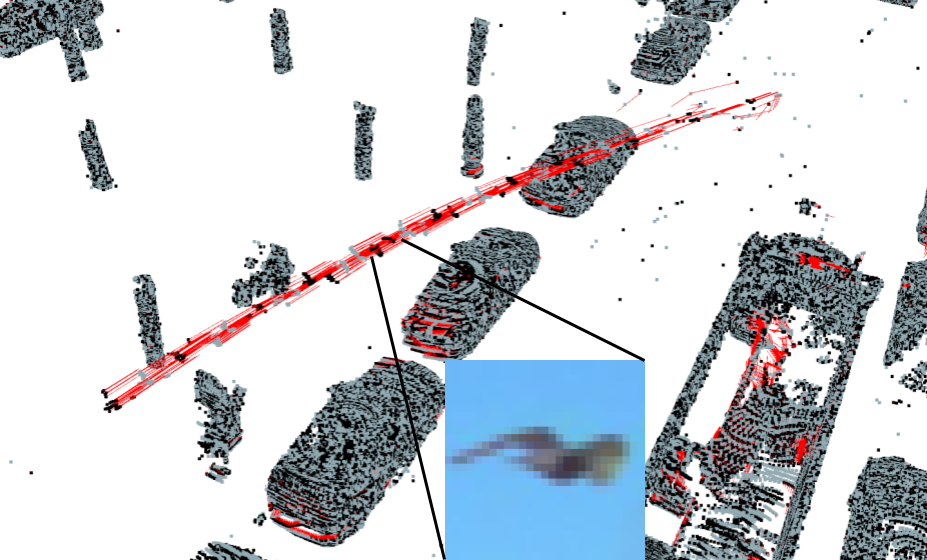}
        \caption{\captionfontsize{} Small object motion extraction...}
        \figlabel{teasersceneflow}
    \end{subfigure}
    \hfill
    \begin{subfigure}[b]{0.248\textwidth}
        \centering
        \includegraphics[width=\textwidth]{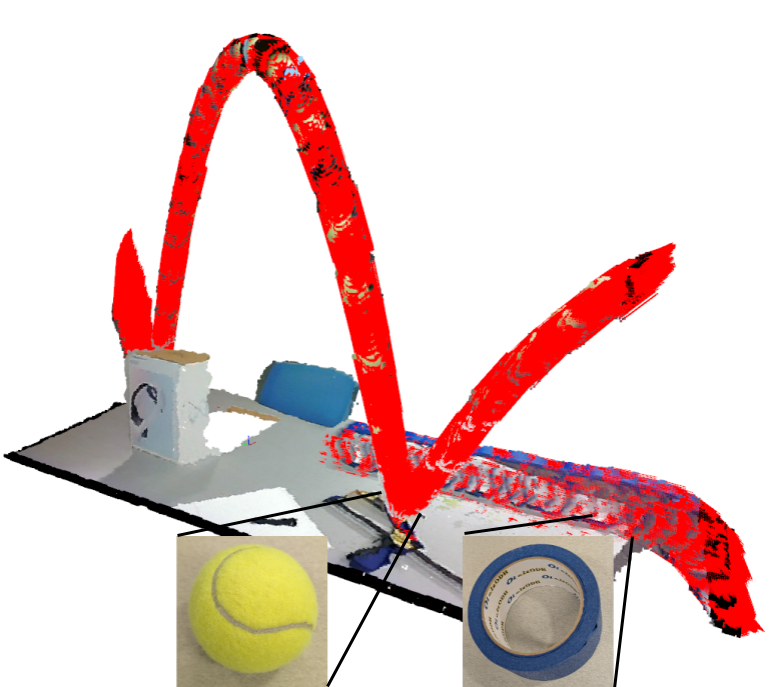}
        \caption{\captionfontsize{} ...in diverse, dynamic scenes...}
        \figlabel{teaserbounce}
    \end{subfigure}
    \hfill
    \begin{subfigure}[b]{0.359\textwidth}
        \centering
        \includegraphics[width=\textwidth]{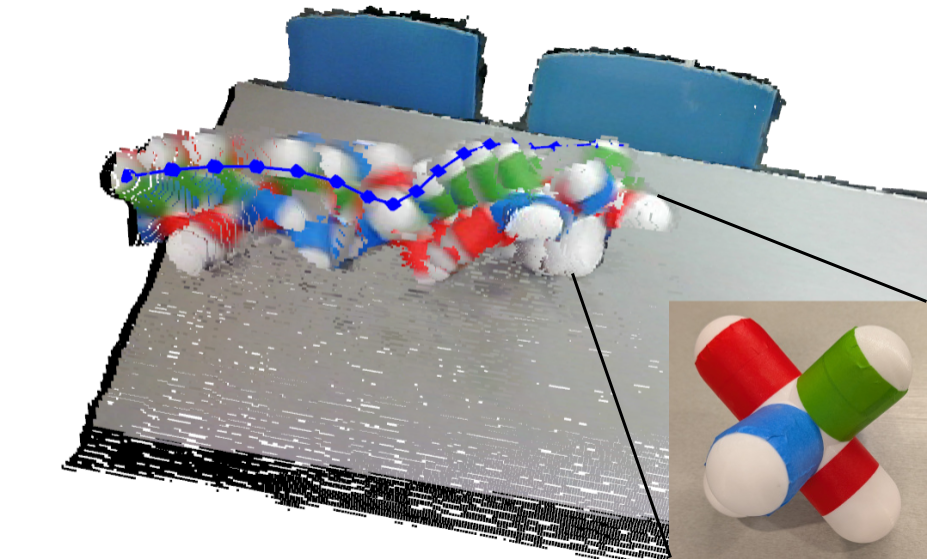}
        \caption{\captionfontsize{} ...with emergent 3D point tracking behavior!}
        \figlabel{teaserjack}
    \end{subfigure}
    \caption{\ourmethod{} is able to capture the motion of small, fast moving objects with few lidar points, such a bird flying in front of an autonomous vehicle (\figref{teasersceneflow}). \ourmethod{}'s flexibility allows it to estimate scene flow for fast-moving table top objects \emph{without additional hyperparameter tuning} (\figref{teaserbounce}). \ourmethod{}'s ODE estimate exhibits emergent 3D point tracking behavior without explicit long-horizon supervision (\figref{teaserjack}). Note that point clouds are shown in color for visualization purposes only; RGB is not used during optimization.}
    \figlabel{teaser}
\end{figure}
%\begin{center}
%Interactive scene visualizations at \websitelink{}
%\end{center}

\begin{abstract}
\input{abstract_hardstyle}
\end{abstract}

\setcounter{footnote}{0} 

\section{Introduction}

Scene flow estimation is the task of describing motion with per-point 3D motion vectors between temporally successive point clouds~\citep{dewan2016rigid,flownet3d, flowssl, scalablesceneflow, seflow,  vedder2024zeroflow, khatri2024trackflow}. Such per-point motion estimates are critical for autonomy in diverse environments, e.g.,\ maneuvering around open-world objects like debris \citep{peri2022towards} or folding deformable cloth \citep{weng2022fabricflownet}. Importantly, scene flow estimation requires not only an understanding of object \emph{geometry}, but also its \emph{motion}. However, scene flow methods broadly do not work on smaller objects \citep{khatri2024trackflow}. For example, in the autonomous vehicles domain, \citeauthor{khatri2024trackflow} highlight that even supervised methods struggle to describe the majority of pedestrian motion, with unsupervised methods failing dramatically. Scene flow promises to be a powerful primitive for understanding the dynamic world, but such failures explain why it has limited adoption in downstream applications like tracking~\citep{Zhai2020FlowMOT3M} or open-world object extraction~\citep{objectdetectionmotion}.

\input{tex_figures/two_frame_vs_full_sequence_2x4}

\poorparagraph{\ourpipelinefull{}} In \figref{walkingpeds}, visual assessment of scene flow quality is much easier in an accumulated global frame; while incomplete due to an implicit time axis, these accumulated flow vectors allow viewers to imagine how positions in 3D space evolve over \emph{many} timesteps, and compare that to predicted flows. This imagination of scene flow as continuous motion over large time intervals motivates us to model scene flow as an ordinary differential equation (ODE) that describes the scene's instantaneous motion across continuous position and time. Scene flow estimation then becomes the task of estimating this ODE. We can straightfowardly represent this ODE estimate with a neural prior~\citep{nsfp} and optimize it against scene flow surrogate objectives, both over single frame pairs and extended \emph{across arbitrary time intervals} to produce better quality estimates. We formalize this in \sectionref{ourpipeline} and propose the \emph{\ourpipelinefull{}} framework.

\poorparagraph{\ourmethod{}} We instantiate \ourpipelinefull{} with standard point cloud distance objectives like Chamfer Distance to create \emph{\ourmethod{}}. Notably, \ourmethod{} outperforms \emph{all} prior art, supervised or unsupervised, on the Argoverse~2 2024 Scene Flow Challenge and Waymo Open Scene Flow benchmark. It outperforms prior \emph{unsupervised} methods by a large margin ($>2.5\times$ mean dynamic error reduction), and is able to capture small, fast moving objects, including those outside of labeled taxonomies (e.g.\ the flying bird in \figref{teasersceneflow}). Due to its simplicity, \ourmethod{} is able to provide high quality scene flow out-of-the-box on real-world data for other important domains such as dynamic tabletop settings (\figref{teaserbounce}) \emph{without} domain-specific tuning. Finally, we show that simple ODE solving techniques like Euler integration can be used to extract 3D point tracks (\figref{teaserjack}), which serves as both an exciting emergent behavior as well as a mechanism for visualizing and interpreting the quality of the continuous ODE estimate.

We present four primary contributions:
\vspace{-1em}
\begin{list}{$\bullet$}{\setlength{\leftmargin}{10pt}} % Adjust leftmargin for indentation
  \setlength{\itemsep}{0pt}   % Adjust space between items
  \setlength{\parskip}{0pt}   % Adjust space between paragraphs
    \item We propose \emph{\ourpipelinefull{}} (\ourpipeline{}), a reframing of scene flow estimation as the task of fitting an ODE that describes the change of continuous positions over continuous time, unlocking a new class of surrogate objectives that enable better scene flow estimates.
    \item We instantiate \ourpipeline{} with \emph{\ourmethod{}}, a flexible \textbf{unsupervised} scene flow method that achieves \textbf{state-of-the-art} performance on the Argoverse~2 2024 Scene Flow Challenge, \textbf{beating all prior supervised and unsupervised methods}.
    \item We study \ourmethod{} and show its strong performance is derived from its ability to optimize its ODE estimate against the full sequence of observations over arbitrary time horizons.
    \item We show that \ourmethod{}'s simple, flexible formulation allows it to run unmodified on a variety of domains, with emergent capabilities like 3D point tracking behavior.
\end{list}

\section{Background and Related Work}\sectionlabel{relatedwork}

\poorparagraph{Evaluation} \citeauthor{dewan2016rigid} formalized scene flow for point clouds as the task of estimating motion between point cloud $\pointcloudt{}$ at time $t$ and point cloud $\pointcloudtpone{}$ at $t+1$ by estimating the true flow $\flowgtttpone$, i.e.\ true residual vectors  for every point in $\pointcloudt{}$ that describe its movement to its associated position at $t+1$. Error is computed by measuring the per-point endpoint distance between estimated and ground truth vectors. Historically, these errors are reported with a per-point average (\emph{Average EPE}); however, as \citeauthor{chodosh2023} show, Average EPE is dominated by background points, preventing meaningful measurement of non-ego object motion descriptions. \citeauthor{khatri2024trackflow} address this shortcoming with \emph{Bucket Normalized EPE}, which reports per-class performance normalized by speed, allowing for direct comparisons across classes with very different average speeds (e.g. pedestrians and cars). Bucket Normalized EPE is the basis for the \emph{Argoverse 2024 Scene Flow Challenge}\footnote{\url{https://www.argoverse.org/sceneflow}}, where methods are ranked by the mean error of their motion descriptions (\emph{mean Dynamic Normalized EPE}).

\poorparagraph{Input / Output Formulation} \citeauthor{dewan2016rigid}'s choice to formulate scene flow using \emph{only} two input frames is arbitrary; it's the minimal information needed to extract rigid motion, but there are not real-world problems constrained to \emph{only} have access to two frames. Indeed, using five or ten frames of past observations is standard practice in the 3D detection literature~\citep{cbgs, vedder2022sparse, peri2022futuredet,peri2023empirical, surveyofpersonrecog}, and multi-frame formulations have started to appear in the scene flow literature: \citet{liu2024selfsupervisedmultiframeneuralscene} and Flow4D~\citep{flow4d} use three ($\pointcloudtmone, \pointcloudt, \pointcloudtpone$) and  five input frames ($\pointcloudsub{t-3}, \ldots, \pointcloudsub{t+1}$) respectively to predict $\flowttpone$. As we discuss in \sectionref{ourpipeline}, rather than just using more observations to estimate flow for a single frame pair, we formulate scene flow as a joint estimation problem: given the full observation sequence $\left( \pointcloud_0, \ldots,\pointcloud_N \right)$, we estimate \emph{all} flows $\flow_{0,1}, \ldots, \flow_{N-1,N}$ between \emph{all} adjacent observations. 

%\neehar{Do we want to reiterate that despite our different input, we're still solving the same 2-frame problem as everyone else? At some other point in the paper, do we need to address that our approach is no longer "online" e.g., we leverage both past and future frames, vs. everyone else only uses past frames?}  % Eulerflow reasons across 10X more frames; other folks would likely run out of memory

\poorparagraph{Feedforward Methods} Feedforward networks are a popular class of scene flow methods due to their fast inference speed~\citep{flownet3d,behl2019pointflownet,tishchenko2020self,kittenplon2021flowstep3d,wu2020pointpwc,puy2020flot,li2021hcrf,scalablesceneflow,gu2019hplflownet,battrawy2022rms, 9856954, flow4d, zhang2024deflow}. While they are often trained with supervised labels, recent work has developed distillation pipelines that leverage unsupervised pseudolabelers~\citep{vedder2024zeroflow,seflow,lin2024icp}. 

%While techniques to handle a variable number of points have been borrowed from the 3D detection literature (e.g.\ \citeauthor{scalablesceneflow} borrow from \citet{pointpillars}; \citeauthor{zhang2024deflow} borrow from \citet{yan2018second}), no techniques are currently used to handle a variable number of pointclouds.

%This full sequence scene flow formulation requires methods that can flexibly handle varying length \emph{sequences} in addition varying numbers of per-frame points. This poses a challenge to feed-forward scene flow methods; while modern architectures can handle variable numbers of points in a point cloud by vectorizing pillars~\citep{scalablesceneflow,zhang2024deflow} or voxels~\citep{flow4d}, the literature does not (yet!) describe any encoding schemes to handle an arbitrary number of distinct point clouds. On the other hand, test-time optimization methods construct their representation on a per-scene basis, enabling a variety of encoding schemes constructed for that particular scene, and thus that scene's sequence length~\citep{pontes2020scene,nsfp,fastnsf}.

\newcommand{\nsfpforward}{\network}
\newcommand{\nsfpbackward}{\network'}

\poorparagraph{Neural Scene Flow Prior} 
\citet{nsfp} propose Neural Scene Flow Prior (NSFP), a widely adopted unsupervised scene flow approach. NSFP uses the inductive bias of the smooth, restricted learnable function class of two ReLU MLP coordinate networks (8 hidden layers of 128 neurons); $\nsfpforward$ to estimate forward flow and $\nsfpbackward$ to estimate backwards flow, minimizing %The first network fits the forward flow $\flowforward$ and the second network fits the backwards flow $\flowrev$, minimizing

\begin{equation}
\small
  \equationlabel{nsfploss}
  \chamferdistance{\pointcloudt{} + \nsfpforward\left(\pointcloudt{}\right)}{\pointcloudtpone} + \norm{\pointcloudt{} + \nsfpforward\left(\pointcloudt{}\right) + \nsfpbackward\left(\pointcloudt{} + \nsfpforward\left(\pointcloudt{}\right) \right) - \pointcloudt}_2 \enspace ,
\end{equation}

where $\chamferdistancename$ is defined as the standard $L_2$ Chamfer distance, but with per-point distances above 2 meters set to zero in order to reduce the influence of outliers. NSFP is optimized for at most 1000 steps with early stopping.

%\citet{liu2024selfsupervisedmultiframeneuralscene} leverages 3 frames of observations to estimate \ ensembles runs of FastNSF (NSFP, but Chamfer Distance is replaced with Distance Transform~\citep{fastnsf}) on $\pointcloudt$ to $\pointcloudtpone$ and a backwards flow from $\pointcloudt$ to $\pointcloudtmone$

% leverages the fact that motion between $\pointcloudtmone$ and $\pointcloudt$ is highly informative of the motion between $\pointcloudt$ and $\pointcloudtpone$ and vice versa by ensembling the flow vectors of multiple runs of FastNSF (NSFP, but Chamfer Distance is replaced with Distance Transform~\citep{fastnsf}) in a two-stage approach. They first use FastNSF to compute a forward flow from $\pointcloudt$ to $\pointcloudtpone$ and a backwards flow from $\pointcloudt$ to $\pointcloudtmone$, resulting in two flow estimates that are ensembled as inputs into a shallow neural prior (4 layer depth) again optimized for the FastNSF loss to produce $\flowttpone$.

\poorparagraph{Motion Beyond Two Frames} 
 \citet{ntp} tackles the adjacent problem of estimating 3D point \emph{trajectories} over 25 frames with Neural Trajectory Prior (NTP) by jointly optimizing three separate ReLU MLP neural priors: 1) a sinusoidal embedded, time conditioned, 25 frame trajectory basis estimator ($\textup{embed}(t) \mapsto 256 \times 25 \times 3$ tensor, where $256$ is the dimension of the trajectory basis), 2) a coordinate network bottleneck encoder, and 3) a bottleneck decoder to estimate a per-point linear combination over the learned trajectories. Trajectories are optimized for both a one-frame lookahead $L_2$ Chamfer loss and a cyclic consistency loss over the entire velocity space trajectory.

\poorparagraph{Deformation in Reconstruction} Nerfies~\citep{park2021nerfies} and DynamicFusion~\citep{dynamicfusion} estimate a deformation field to warp a canonical frame to explain the observed frame. While capable of describing small motions, these methods require a canonical frame that contains all of the relevant geometry to deform; however, in large, highly dynamic scenes like autonomous driving, there is often no frame that contains all moving objects. By comparison, \ourpipelinefull{} does not assume the existence of a canonical frame, instead only describing how the scene changes.

\begin{figure}[tbh]
    \centering
%https://docs.google.com/drawings/d/1idlhkbFNxbIiy5My2uQih_-JzcUp8sXiNb1ZEyWisy4/edit?usp=sharing
\includegraphics{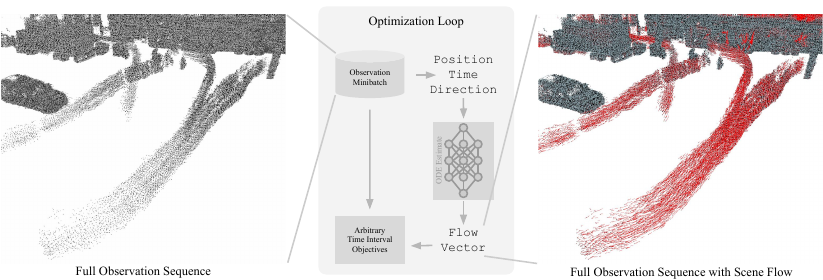}
    \caption{Overview of our \emph{\ourpipelinefull{}} framework, which estimates an ODE across the entire observation sequence by optimizing against multi-frame objectives. This ODE estimate is represented with a neural prior \citep{nsfp}, providing a flexible, general representation for describing position-time motion.\vspace{-1em}}
    \figlabel{method}
\end{figure}

\section{\ourpipelinefull{}}\sectionlabel{ourpipeline}

Prior art consumes multiple frames $\left(\pointcloud_{t-N}, \ldots, \pointcloudtpone\right)$ as input, but these methods are ultimately only tasked with estimating flow vectors between $\pointcloudt$ and $\pointcloudtpone$. We instead pose the problem of estimating a time-conditioned flow field that describes motion for \emph{all} adjacent point clouds $\pointcloudt, \pointcloudtpone$ in the entire sequence $\left( \pointcloud_0, \ldots,\pointcloud_N \right)$. To do this, rather than describing scene flow as positional change over a fixed interval ($\flowgtttpone$ are residual vectors over the interval $t$ to $t+1$) as we did in \sectionref{relatedwork}, we can instead express these changes as a differential equation that describes positional change over \emph{continuous} time.

\input{tex_figures/field_descriptions}

Formally, given a scene, let $L(x_0,y_0,z_0,t)$ be the Lagrangian view of the scene's true flow field, i.e.\ a continuous function that, based on a canonical frame at time $0$, describes the true position of the canonical frame particle $x_0,y_0,z_0$ at some other time $t$. As we discuss in \sectionref{relatedwork}, this Lagrangian view is common in the the deformable reconstruction literature, and the requirement for a canonical frame definition means these approaches struggle to describe scenes where there is no frame that contains all moving objects.

To break this canonical frame dependence, we choose to take an Eulerian view of the flow field, i.e.\ $F = \frac{d L}{d t}$, which describes the velocity of a query point at some arbitrary time. As we show in our derivation in \appendixref{derivation}, this formulation does not require point correspondences in some other canonical frame when estimating a point's trajectory from $t$ to $t'$; instead, we can simply set the initial conditions of the ODE at $t$ to $x_t, y_t, z_t$ and utilize an off-the-shelf ODE solver (e.g. Euler integration) to extract flow from $t$ to $t'$, expressed as $E(x_t, y_t, z_t, t, t')$.

We do not know the true flow field $F$ when estimating scene flow; however, we can represent $F$ with a neural prior $\network$ ($F \approx \network$), and optimize $\network$ against surrogate objectives. This framing, which we formalize into the \emph{\ourpipelinefull{}}  framework (\ourpipeline{}; \figref{method}), allows $\network$ to benefit from constructive interference between objectives, as well as enables us to formulate objectives over arbitrarily long time horizons, unlocking high quality estimates.

\section{\ourmethod{}}

\emph{\ourpipelinefull{}} proposes a framework where the neural prior $\network$ represents an estimate of the Eulerian flow field $F$  (i.e. $F \approx \network$); however, it does not prescribe the optimization objectives for $\network$. Thus, we instantiate {\ourpipelinefull{}} with \emph{\ourmethod{}}, a point cloud only scene flow method\footnote{Visualizations shown in color for better viewing. 
\ourmethod{} can also use monodepth estimates (\appendixref{monodepth})} with reconstruction and cyclic consistency objectives across the entire sequence of observations.

As we show in \equationref{eulertrajectory} (\appendixref{neuralpriorrollout}), we can use $\network$'s Eulerian flow field estimate to extract an estimated point trajectory from $x_t, y_t, z_t$ at $t$ to some future location at time $t'$ via Euler integration over $\network$ without requiring a canonical frame definition, i.e. $E_\network(x_t, y_t, z_t, t, t')$. By extracting point trajectories for every point $p$ in $\pointcloudt$ using $E_\network$, we can not only construct a two-frame scene flow estimate of $\flowgtttpone$, but also estimate flow to arbitrary future or prior timesteps (e.g.\ $\flowgt_{t,t+2}$ or $\flowgt_{t, t-1}$). This allows us to optimize over multi-frame reconstruction objectives: we can now pose reconstruction surrogate objectives between \emph{any} two point clouds in our observation sequence, not just adjacent point clouds $\pointcloudt$ and $\pointcloudtpone$. Similarly, we can straightforwardly pose cyclic consistency objectives by composing $\flowgt_{t,t+1}$ and $\flowgt_{t + 1, t}$. Formally, for $\pointcloudt$'s $\flowgt_{t, t+k}$ (for any $k \in \mathbb{Z}$), we define

\begin{equation}
\kstep{\pointcloudt}{k} = \pointcloudt + \flowgt_{t, t+k} = \forall p \in \pointcloudt: E_\network(p_{xt}, p_{yt}, p_{zt}, t, t+k)\enspace ,
\end{equation}

enabling us to pose $\network$'s optimization objective $\forall \pointcloudt \in \left( \pointcloud_0, \ldots,\pointcloud_N \right)$ across the window of size $W$

\begin{equation}\equationlabel{gigachadloss}
    \arg\min_{\network} \sum{
    \begin{array}{l}
        \forall k \in \{-W, \ldots, W\} \setminus \{0\}:  \chamferdistance{\kstep{\pointcloudt}{k}}{\pointcloudsub{t+k}}\\
        \alpha \norm{\kstep{\kstep{\pointcloudt}{1}}{-1} - \pointcloudt}_2
        % \norm{\fwdcycle{\pointcloudt} - \pointcloudt}_2 \times 0.01
        % 
    \end{array}
    } \enspace
\end{equation}

In practice, we set W to 3 and $\alpha$ to $0.01$. We provide additional implementation details in \appendixref{implementationdetails}. In order to optimize $\network$, our estimate of the Eulerian flow field $F$, we perform Euler integration to extract point cloud flow estimates as part of reconstruction losses. Notably, \ourmethod{} only requires a single optimization loop over a single neural prior $\network$ compared to NSFP's two priors $\network$ and $\network'$. Our neural prior $\network$ is a straightforward extension to NSFP's coordinate network prior. Like with NSFP, $\chamferdistancename$ is defined as the standard $L_2$ Chamfer distance with per-point distances below 2 meters. As we show in \sectionref{experiments}, \ourmethod{}'s simple ODE estimation formulation across multiple observations produces high quality flow, and solving this ODE over arbitrary time spans unlocks emergent point tracking behavior. 

\section{Experiments}\sectionlabel{experiments}

In order to validate \ourmethod{}'s construction and better understand the impact of its design choices, we perform extensive experiments on the Argoverse~2~\citep{argoverse2} and Waymo Open~\citep{waymoopen} autonomous vehicle datasets. We compare against open source implementations of FastNSF~\citep{fastnsf}, \citeauthor{liu2024selfsupervisedmultiframeneuralscene}, NSFP~\citep{nsfp}, FastFlow3D~\citep{scalablesceneflow}, and variants of ZeroFlow~\citep{vedder2024zeroflow} provided by the ZeroFlow model zoo\footnote{\url{https://github.com/kylevedder/SceneFlowZoo}, from \citet{vedder2024zeroflow}.}, a third-party implementation of NTP~\citep{ntp} from \citeauthor{vidanapathirana2023mbnsf}, and Argoverse 2 2024 Scene Flow Challenge leaderboard submission results from the authors of Flow4D~\citep{flow4d}, TrackFlow~\citep{khatri2024trackflow}, DeFlow++/DeFlow~\citep{zhang2024deflow}, ICP Flow~\citep{lin2024icp}, and SeFlow~\citep{seflow}. As discussed in \citeauthor{khatri2024trackflow} and used in the Argoverse~2 2024 Scene Flow Challenge, methods are ranked by their speed normalized \emph{mean Dynamic Normalized EPE}. 

\poorparagraph{Implementation Details} To showcase the flexibility of \ourmethod{} without hyperparameter tuning, for all quantitative experiments we run with a neural prior of depth 8 (NSFP's default depth), except for our submission to the Argoverse 2 2024 Scene Flow Challenge (\sectionref{performance}) where, based on our depth ablation study on the val split (\sectionref{mlpdepth}), we set the depth of the neural prior to 18. As discussed in NTP's original paper \citep{ntp} and confirmed by our experiments, NTP struggles to converge beyond 25 frames, so we only compare against it in a 20 frame settings. As is typical in the scene flow literature~\citep{chodosh2023}, we perform ego compensation and ground point removal on both Argoverse~2 and Waymo Open using the dataset provided map and ego pose.

\subsection{How does \ourmethod{} compare to prior art on real data?}\sectionlabel{performance}
\ourmethod{} achieves \textbf{state-of-the-art} performance on the \emph{Argoverse 2 2024 Scene Flow Challenge} leaderboard.
Despite being unsupervised, \ourmethod{} \textbf{surpasses \emph{all} prior art, supervised or unsupervised}, including Flow4D~\citep{flow4d}\footnote{Flow4D is the winner of the 2024 Argoverse 2 Scene Flow Challenge supervised track.} and ICP~Flow~\citep{lin2024icp}\footnote{ICP~Flow is the winner of the 2024 Argoverse 2 Scene Flow Challenge unsupervised track.}. Notably, \ourmethod{} achieves $< 2.5\times$ lower error mean Dynamic EPE than ICP~Flow and beats Flow4D by over 10\%.

\begin{figure}[htb]
    \centering
    \includegraphics{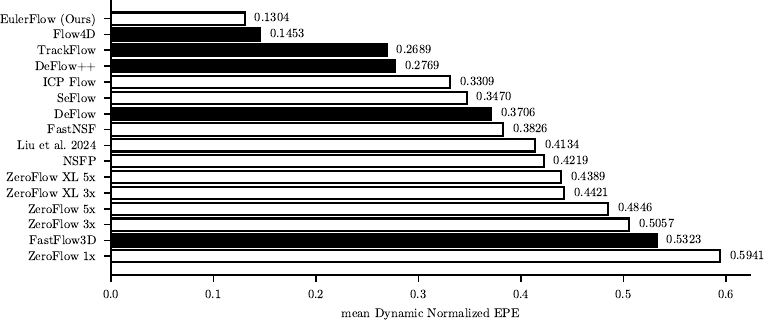}
    \caption{Mean Dynamic Normalized EPE of \ourmethod{} compared to prior art on the Argoverse~2 2024 Scene Flow Challenge test set. \ourmethod{} is state-of-the-art, beating all supervised (shown in black) and unsupervised (shown in white) methods. Lower is better.\vspace{-1em}}
    \figlabel{argodynamicepe}
\end{figure}

\begin{figure}[htb]
    \centering
    \includegraphics{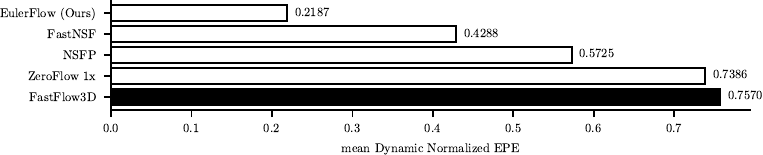}
    \caption{Mean Dynamic Normalized EPE of \ourmethod{} compared to prior art on the Waymo Open validation set. \ourmethod{} is state-of-the-art, beating all supervised (shown in black) and unsupervised (shown in white) methods. Lower is better.\vspace{-1em}}
    \figlabel{waymodynamicepe}
\end{figure}

\ourmethod{}'s dominant performance also holds on Waymo Open~\citep{waymoopen}; we compare against several popular methods (\figref{waymodynamicepe}), and \ourmethod{} again out-performs the baselines by a wide margin, more than halving the error over the next best method. 

\subsection{What contributes to \ourmethod{}'s state-of-the-art performance?}

\begin{figure}[htb]
\centering
% First row
\begin{subfigure}[b]{0.49\textwidth}
    \centering
    \includegraphics{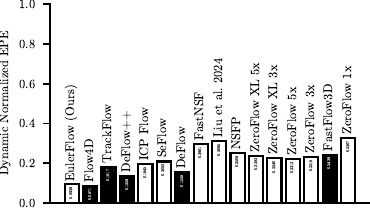}
    \caption{\texttt{CAR}}
    \figlabel{fig:car}
\end{subfigure}%
\begin{subfigure}[b]{0.49\textwidth}
    \centering
    \includegraphics{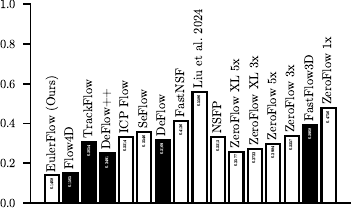}
    \caption{\texttt{OTHER VEHICLES}}
    \figlabel{fig:other-vehicles}
\end{subfigure}
% Second row
\begin{subfigure}[b]{0.49\textwidth}
    \centering
    \includegraphics{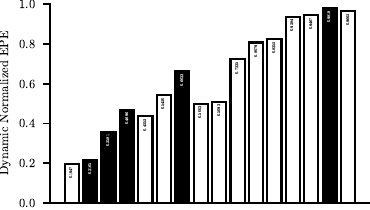}
    \caption{\texttt{PEDESTRIAN}}
    \figlabel{fig:pedestrian}
\end{subfigure}%
\begin{subfigure}[b]{0.49\textwidth}
    \centering
    \includegraphics{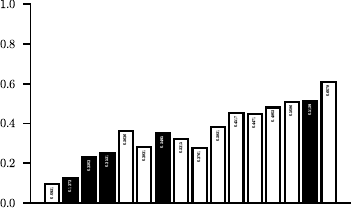}
    \caption{\texttt{WHEELED VRU}}
    \figlabel{fig:wheeled-vru}
\end{subfigure}
\caption{Per class Dynamic Normalized EPE of \ourmethod{} compared to prior art on the Argoverse~2 2024 Scene Flow Challenge test set. Supervised methods shown in black, unsupervised methods shown in white. Methods are ordered left to right by increasing mean Dynamic Normalized EPE. Lower is better. \vspace{-1em}}
\figlabel{argometacatagorydynamic}
\end{figure}

\begin{figure}[h!]
\centering
% First row
\begin{subfigure}[b]{0.32\textwidth}
    \centering
    \includegraphics{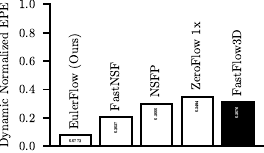}
    \caption{\texttt{VEHICLE}}
    \figlabel{fig:waymovehicle}
\end{subfigure}%
\begin{subfigure}[b]{0.32\textwidth}
    \centering
    \includegraphics{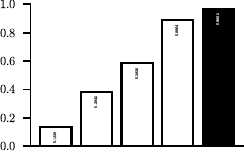}
    \caption{\texttt{CYCLIST}}
    \figlabel{fig:waynocyclist}
\end{subfigure}%
\begin{subfigure}[b]{0.32\textwidth}
    \centering
    \includegraphics{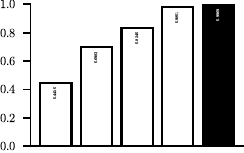}
    \caption{\texttt{PEDESTRIAN}}
    \figlabel{fig:waymopedestrian}
\end{subfigure}
\caption{Per class Dynamic Normalized EPE of \ourmethod{} compared to prior art on the Waymo Open validation set. Supervised methods shown in black, unsupervised methods shown in white. Methods are ordered left to right by increasing mean Dynamic Normalized EPE. Lower is better.\vspace{-1em}}
\figlabel{waymometacatagorydynamic}
\end{figure}

We find that \ourmethod{}'s lower mean Dynamic EPE can be attributed to better performance on smaller objects. On Argoverse~2, compared to Flow4D, \ourmethod{}'s improves on \texttt{WHEELED VRU} (\figref{fig:wheeled-vru}), a small, rare, fast moving class. Compared to ICP~Flow, \ourmethod{}'s improves on all classes (at least halving the error on every class!), with the largest improvements coming from the smaller and harder to detect objects \texttt{PEDESTRAIN} and \texttt{WHEELED VRU} (\figrefs{fig:pedestrian}{fig:wheeled-vru}). On Waymo Open, the same story holds; the most dramatic performance improvements come from the small object classes of \texttt{CYCLIST} and \texttt{PEDESTRIAN} (\figref{waymometacatagorydynamic}).

These results are consistent with our qualitative visualizations. \figref{flyingbird} shows \ourmethod{} is able to cleanly extract the motion of a bird flying past the ego vehicle. Euler integration using \ourmethod{}'s ODE, starting at the bird's takeoff position and ending when it loses lidar returns, produces emergent 3D point tracking behavior on the bird through its trajectory (\figref{birdtracking}), further demonstrating the quality of \ourmethod{}'s model of the scene's motion. 

\begin{figure}[htb]
\centering
% First row
\begin{subfigure}[b]{0.49\textwidth}
    \centering
    \includegraphics[width=\textwidth]{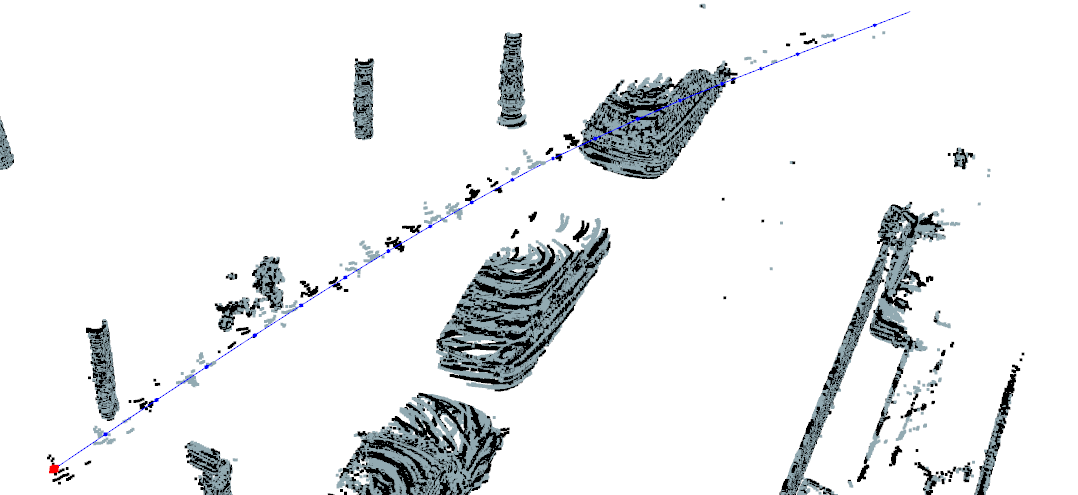}
    \caption{Bird trajectory via Euler integration from takeoff}
    \figlabel{birdtrajectory}
\end{subfigure}%
\begin{subfigure}[b]{0.49\textwidth}
    \centering
    \includegraphics[width=\textwidth]{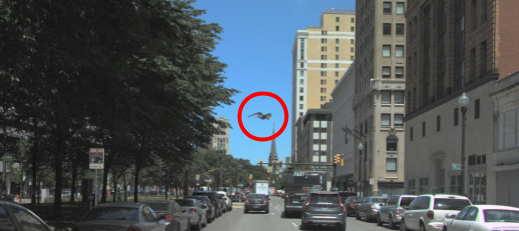}
    \caption{Bird being tracked}
    \figlabel{birdrgb}
\end{subfigure}
    \caption{\ourmethod{} is able to track the bird over 20 frames by performing Euler integration starting from takeoff until it loses all point cloud lidar returns.}
    \figlabel{birdtracking}
\end{figure}

% \subsection{What are the keys to \ourmethod{}'s success?}\sectionlabel{gigachadablations}

% \ourmethod{}'s dominant performance raises the question: what are the key ingredients in its success? 

\subsubsection{How does observation sequence length impact \ourmethod{}?}\sectionlabel{sequencelength}

 \begin{figure}[htb]
     \centering
    \includegraphics{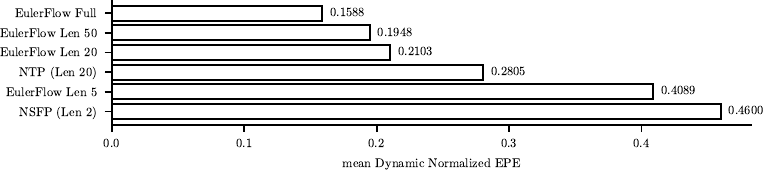}
     \caption{Mean Dynamic Normalized EPE of \ourmethod{} for various sequence lengths on the Argoverse 2 val split, compared against representative baselines. These results demonstrate that \ourmethod{} improves with sequence length; however, at a sequence length of 20, our method significantly outperforms NTP, suggesting that \ourmethod{}'s performance cannot solely be attributed to longer sequence modeling. \vspace{-1em}}
     \figlabel{sequencelength}
 \end{figure}

As we discuss in \sectionref{ourpipeline}, \ourmethod{} benefits from constructive interference from ODE estimation over many observations. Does this sufficiently explain \ourmethod{}'s performance?
\figref{sequencelength} shows the performance of \ourmethod{} at length 5, 20, 50, and full sequence (roughly 160 frames) compared to NSFP and NTP at length 20. \ourmethod{} sees clear continual improvements as the number of frames increases without signs of saturation. However, sequence length alone does not explain \ourmethod{}'s performance; even at the same sequence length of 20, \ourmethod{} demonstrates significantly better performance than NTP.

\subsubsection{How do multi-frame optimization objectives impact \ourmethod{}?}

\begin{figure}[htb]
    \centering
    \includegraphics{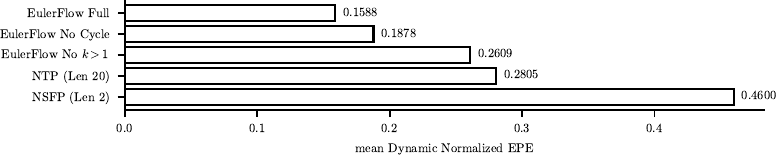}
    \caption{Mean Dynamic Normalized EPE of \ourmethod{} for various losses on the Argoverse 2 val split, compared against representative baselines. These results demonstrate that \ourmethod{}'s multi-observation optimization objectives significantly improve overall performance.\vspace{-1em}}
    \figlabel{lossablations}
\end{figure}

\equationref{gigachadloss} outlines two major components of \ourmethod{}'s loss: multi-frame Euler integration for Chamfer Distance reconstruction, and cycle consistency. \figref{lossablations} compares \ourmethod{} without more than one integration step (No $k>1$) and without cycle consistency rollouts (No Cycle) to better understand the impact of these components. Ablating multi-step Euler integrated rollouts results in significant degredation, as they are a strong forcing function to have consistent, smooth flow volumes; indeed, despite consuming the entire sequence, \ourmethod{} (No $k>1$) is only slightly better than NTP with a sequence length of 20. These results highlight the power of multi-step rollouts and their potential as a objective for other test-time optimization methods and feedforward methods.

\subsubsection{How does the capacity of the neural prior impact \ourmethod{}?}\sectionlabel{mlpdepth}

\citeauthor{nsfp} ablate the capacity of  NSFP's neural prior to characterize underfitting and overfitting to optimization objective noise, ultimately selecting a depth of 8. \ourmethod{}'s neural prior is structured similarly; however, NSFP is fitting a single snapshot in time, while \ourmethod{} is fitting an entire ODE over significant time intervals. Intuitively, one would expect that full sequence modeling would benefit from greater network capacity.

To evaluate this, we perform a sweep of \ourmethod{}'s network depth on the Argoverse 2 validation split (\figref{depthablation}). While \ourmethod{} with NSFP's default of depth 8 performs well on our Argoverse 2 evaluations (0.1\% worse than the supervised state-of-the-art Flow4D), we see that performance improves as the neural prior's depth increases until depth 18 (indicating underfitting), where we start to see degradation (indicating overfitting to noise). Based on these results our Argoverse 2 2024 Scene Flow Challenge leaderboard submission uses a depth 18 neural prior (\figref{argodynamicepe}).

\begin{figure}[htb]
    \centering
    \includegraphics{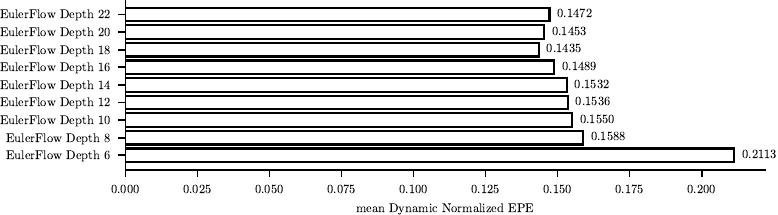}
    \caption{Mean Dynamic Normalized EPE of \ourmethod{} on the Argoverse 2 val split for different neural prior capacities. Shallow networks underfit the ODE, while deeper networks overfit to noise in the optimization objectives.\vspace{-1em}}
    \figlabel{depthablation}
\end{figure}

\input{tex_figures/bird_two_rows}

\subsection{Beyond Autonomous Vehicles}

Due to a dearth of real-world, labeled scene flow data, prior scene flow work on real data overwhelmingly evaluates on autonomous vehicle datasets~\citep{dewan2016rigid, nsfp, scalablesceneflow, fastnsf, chodosh2023,   liu2024selfsupervisedmultiframeneuralscene, vedder2024zeroflow, khatri2024trackflow}; consequently, motion understanding in other important domains like tabletop manipulation has been neglected. To showcase \ourmethod{}'s out-of-the-box flexibility and generalizability, we visualize \ourmethod{} on several dynamic tabletop scenes we collected using the ORBBEC Astra, a low cost depth camera commonly used in robotics (\figref{orbbec}). For viewing ease, we paint our point clouds with color; however, RGB information is not provided to \ourmethod{} during optimization. While \ourmethod{} only reasons about point clouds, it can leverage video mono depth estimates to describe RGB-only scene flow (\appendixref{monodepth}). Interactive visuals are available at \websitelink{}.

\begin{figure}[htb]
    \centering

    \begin{subfigure}[b]{0.24\textwidth}
    \centering
    \includegraphics[width=\textwidth]{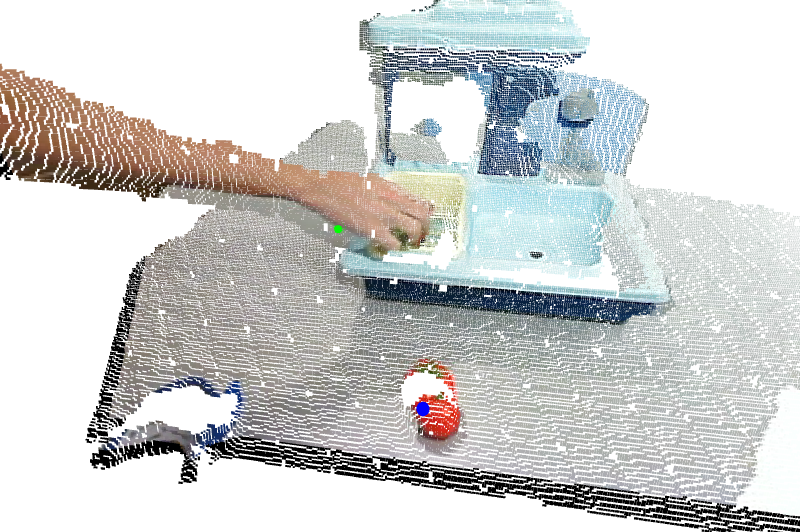}
\end{subfigure}%
\begin{subfigure}[b]{0.24\textwidth}
    \centering
    \includegraphics[width=\textwidth]{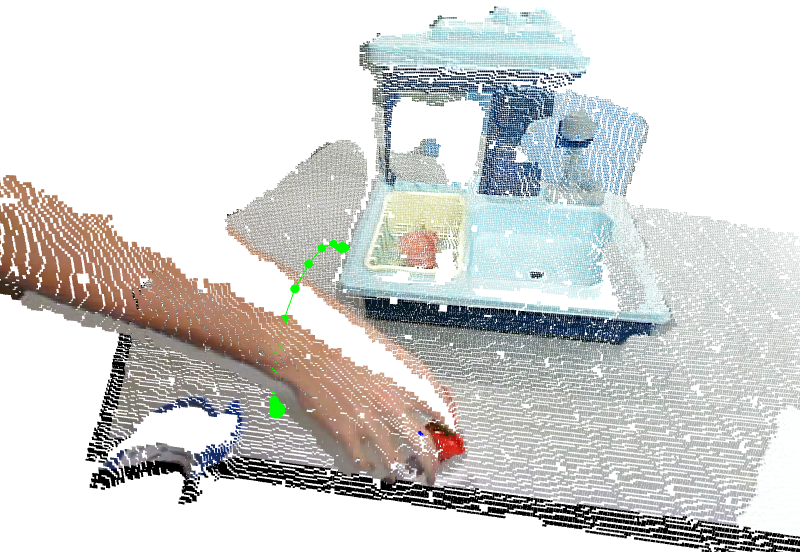}
\end{subfigure}%
\begin{subfigure}[b]{0.24\textwidth}
    \centering
    \includegraphics[width=\textwidth]{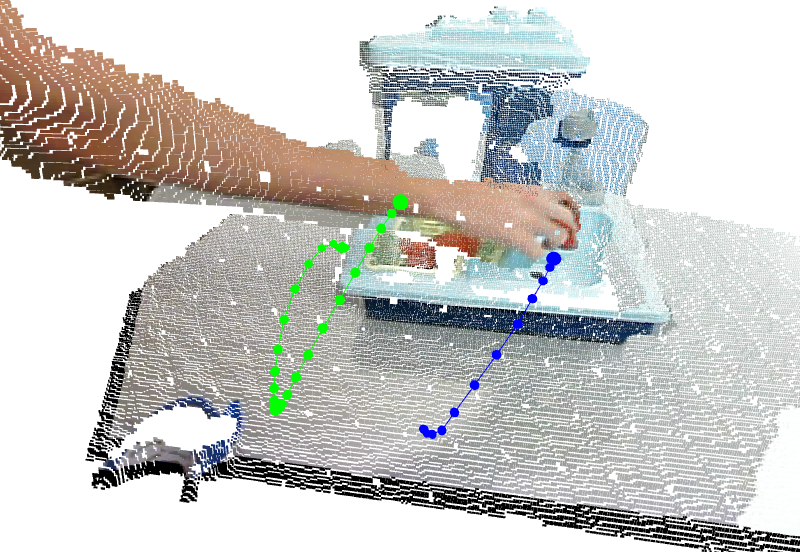}
\end{subfigure}%
\begin{subfigure}[b]{0.24\textwidth}
    \centering
    \includegraphics[width=\textwidth]{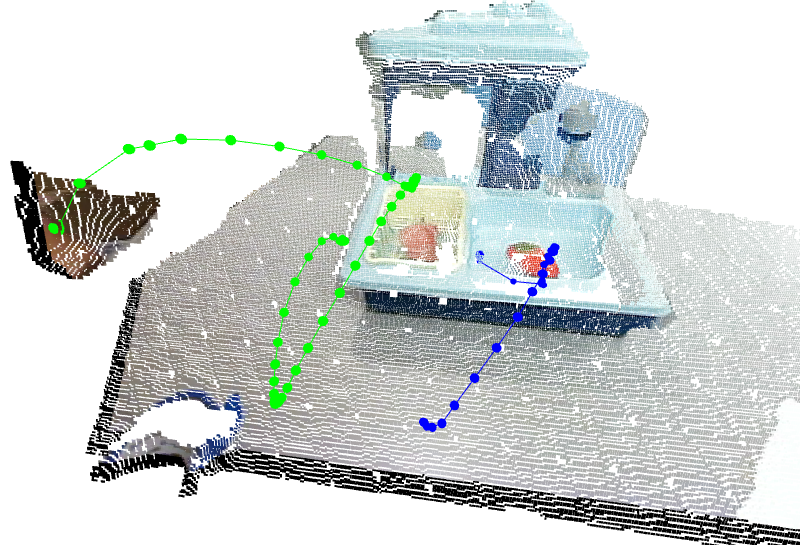}
\end{subfigure}

    \begin{subfigure}[b]{0.24\textwidth}
    \centering
    \includegraphics[width=\textwidth]{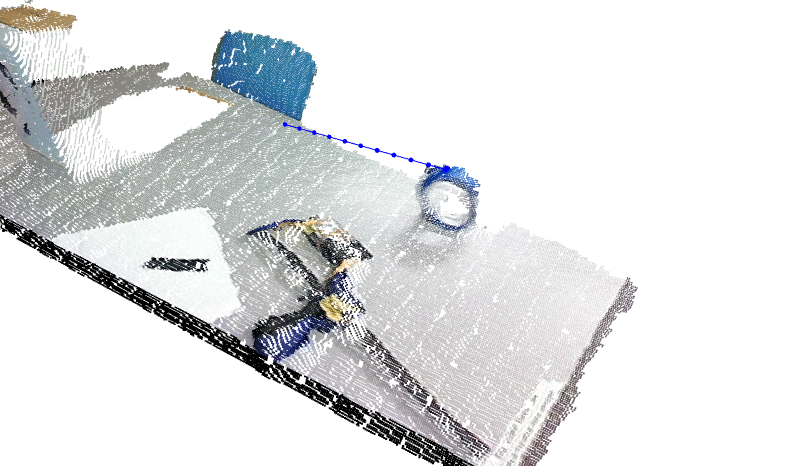}
\end{subfigure}%
\begin{subfigure}[b]{0.24\textwidth}
    \centering
    \includegraphics[width=\textwidth]{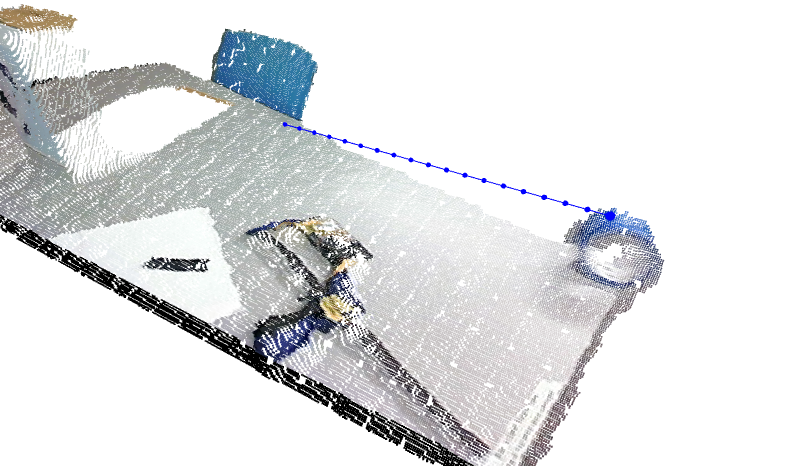}
\end{subfigure}%
\begin{subfigure}[b]{0.24\textwidth}
    \centering
    \includegraphics[width=\textwidth]{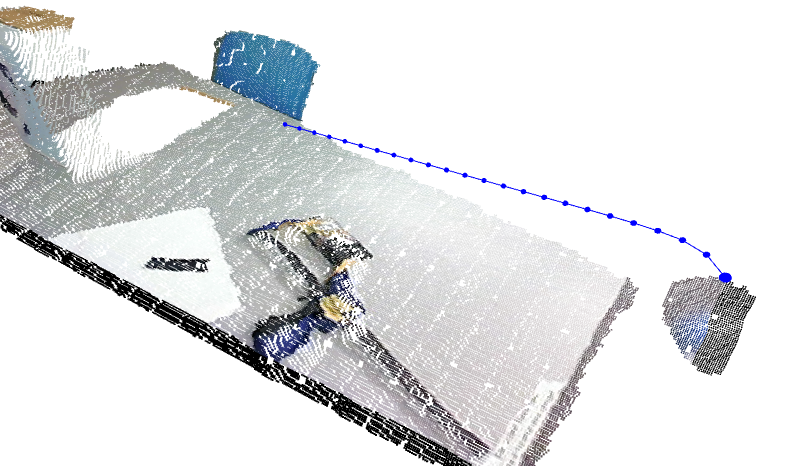}
\end{subfigure}%
\begin{subfigure}[b]{0.24\textwidth}
    \centering
    \includegraphics[width=\textwidth]{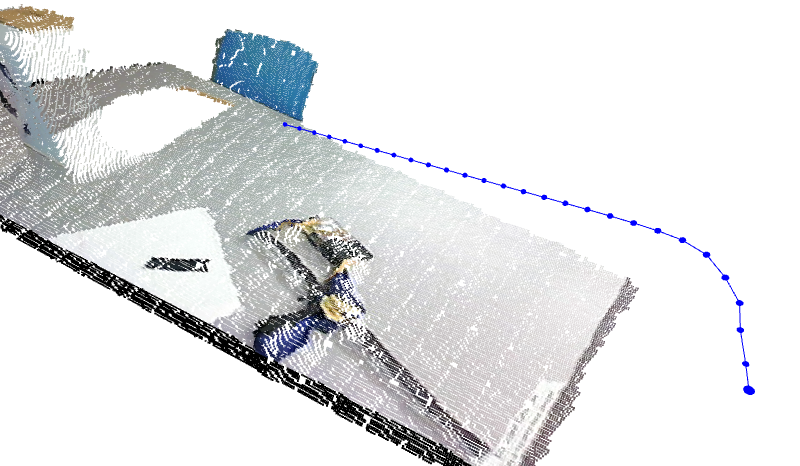}
\end{subfigure}

    \begin{subfigure}[b]{0.24\textwidth}
    \centering
    \includegraphics[width=\textwidth]{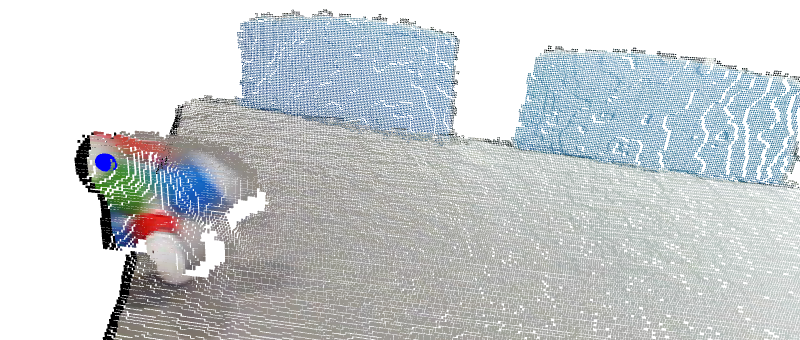}
\end{subfigure}%
\begin{subfigure}[b]{0.24\textwidth}
    \centering
    \includegraphics[width=\textwidth]{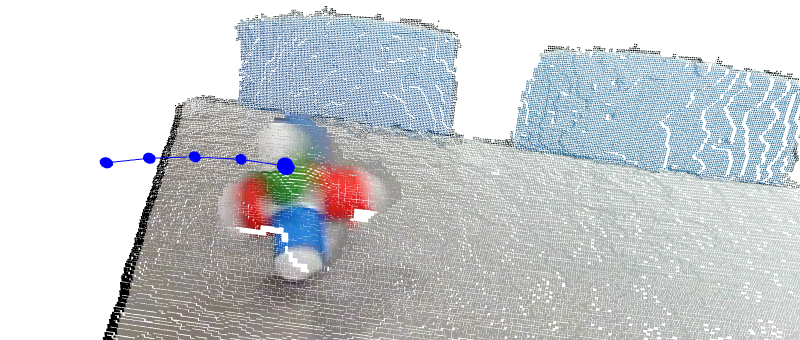}
\end{subfigure}%
\begin{subfigure}[b]{0.24\textwidth}
    \centering
    \includegraphics[width=\textwidth]{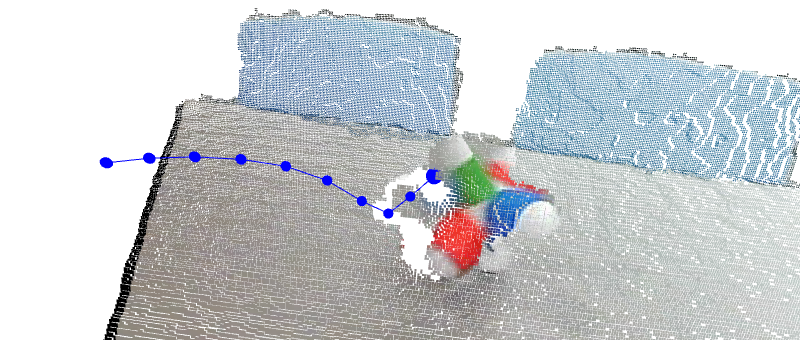}
\end{subfigure}%
\begin{subfigure}[b]{0.24\textwidth}
    \centering
    \includegraphics[width=\textwidth]{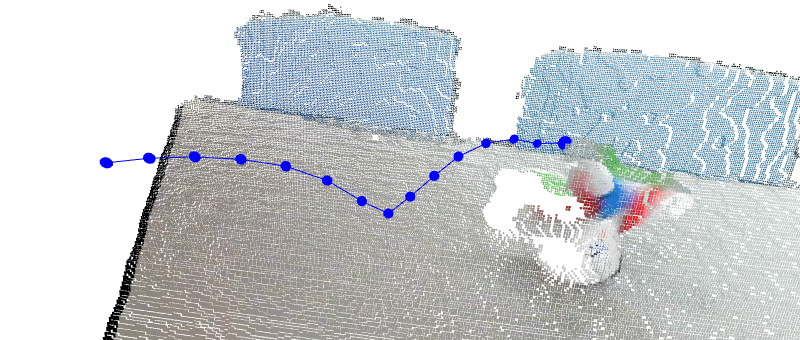}
\end{subfigure}

    \begin{subfigure}[b]{0.24\textwidth}
    \centering
    \includegraphics[width=\textwidth]{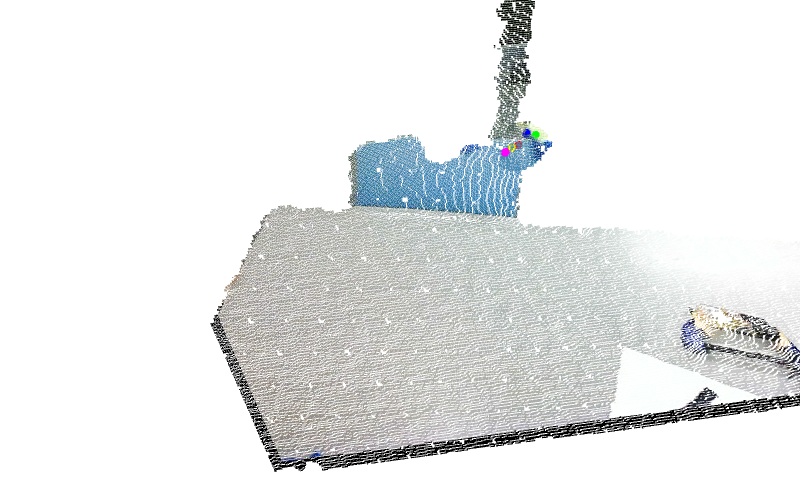}
\end{subfigure}%
\begin{subfigure}[b]{0.24\textwidth}
    \centering
    \includegraphics[width=\textwidth]{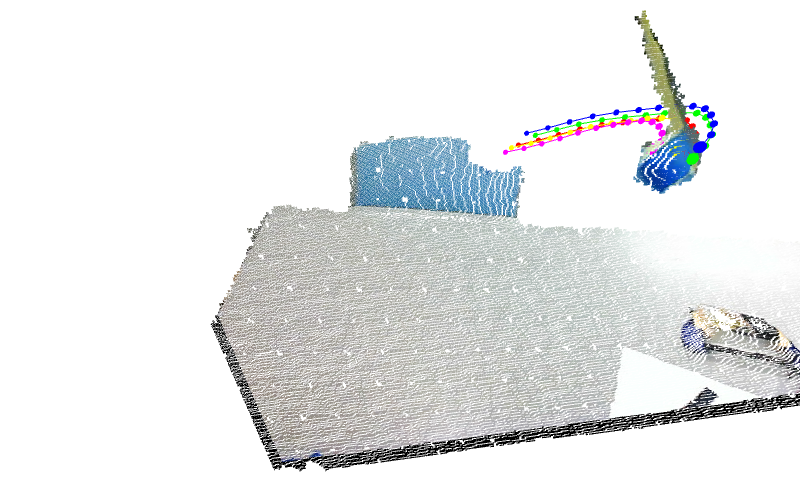}
\end{subfigure}%
\begin{subfigure}[b]{0.24\textwidth}
    \centering
    \includegraphics[width=\textwidth]{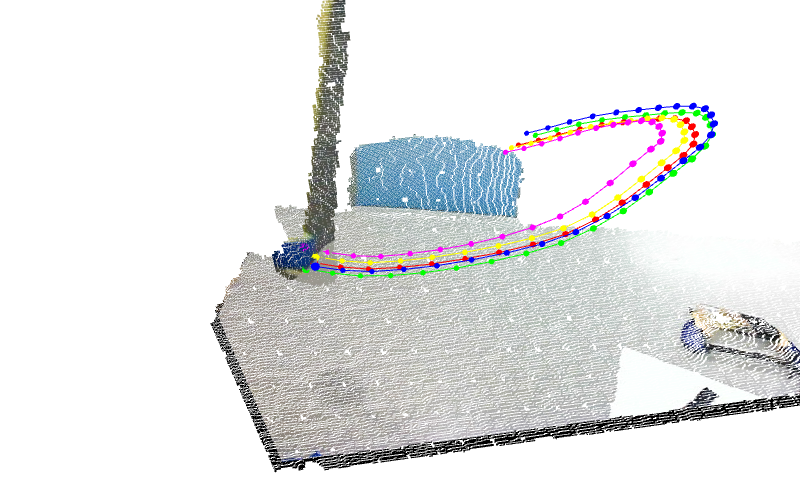}
\end{subfigure}%
\begin{subfigure}[b]{0.24\textwidth}
    \centering
    \includegraphics[width=\textwidth]{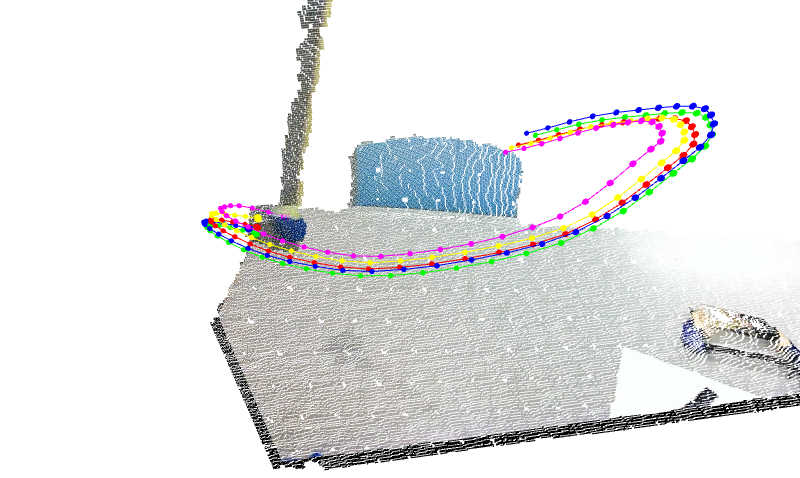}
\end{subfigure}

    \caption{Visualizations of \ourmethod{}'s emergent 3D point tracking behavior that demonstrate the quality of its ODE estimate. Row 1 depicts tracking a tomato placed in the sink by a human hand; note the point does not move despite the hand grasping the tomato. Row 2 depicts tracking of painters tape rolling off a table; \ourmethod{} is able to estimate its trajectory even after it disappears out of frame. Row 3 depicts tracking of the motion of a jack commonly used in tabletop manipulation experiments~\citep{venkatesh2023samplingbased}. Row 4 depicts tracking of a tennis ball taped to a flexible rod. All tracks are produced by Euler integration through the estimated ODE from the initial conditions shown in the left column. Note that point clouds are shown in color for visualization purposes only.\vspace{-1.5em}}
    \figlabel{orbbec}
\end{figure}

\vspace{-0.5em}

\section{Conclusion}

By reframing scene flow as fitting an ODE over positions for a full sequence of observations, we are able to construct \ourmethod{}, a simple unsupervised scene flow method that achieves state-of-the-art performance on the Argoverse 2 2024 Scene Flow Challenge and Waymo Scene Flow benchmark, where it beats all prior art, supervised or unsupervised. \ourmethod{} is able to describe motion on small, fast moving, out of distribution objects unable to be captured by prior art, suggesting that it makes good on the promises of scene flow as a powerful primitive for understanding the dynamic world. It also exhibits other emergent capabilities, like basic 3D point tracking behavior.

We believe that this ODE formulation has implications for scene flow at large, including beyond test-time optimization methods; the power of multi-step Euler integration may translate to feedforward network training. Future work should explore feedforward models that perform autoregressive rollouts or directly learn to estimate multiple steps into the future.

\vspace{-0.5em}
\subsection{Limitations and Future Work}\sectionlabel{limitations}

\ourmethod{}'s strong performance opens the book on an exciting new line of work; however, we feel that it's important to be candid about \ourmethod{}'s current limitations in order to make future progress.

\vspace{-0.2em}
\emph{\ourmethod{} is point cloud only.} Point cloud sparsity bottlenecks performance; for instance, in \figref{birdtracking} and \figref{flyingbird} we were only able to track the bird for 20 frames because we lost lidar observations of the bird, while it remained visible in the car's RGB cameras. Future works should explore multi-modal fusion for better long-term motion descriptions.

\vspace{-0.2em}
\emph{\ourmethod{} is expensive to optimize.} With our implementation, optimizing \ourmethod{} for a single Argoverse 2 sequence takes 24 hours on one NVIDIA V100 16GB GPU, putting it on par with the original NeRF paper's computation expense~\citep{nerf}. However, like with NeRF, we believe  algorithmic, optimization, and engineering improvements can significantly reduce runtime.

\vspace{-0.2em}
\emph{\ourmethod{} does not understand ray casting geometry.} During ego-motion, a static foreground occluding object casts a moving shadow on the background; this causes Chamfer Distance to estimate this as a leading edge of moving structure, encouraging false motion artifacts~\citep{nsfp}. This can be addressed with optimization losses that model point clouds as originating from a time of flight sensor with limited visibility, as has been successfully demonstrated in the reconstruction~\citep{chodosh2024simultaneousmapobjectreconstruction} and forecasting literature~\citep{khurana2023point, agro2024uno}, rather than an unstructured set of points to be associated via local point distance.\vspace{-1em}

\bibliography{references}
\bibliographystyle{iclr2024_conference}

\newpage 

\appendix
\section{Additional results}

\subsection{How does the choice of learnable function class and design of encodings impact \ourmethod{}?}

\ourmethod{} at its core is an optimization loop over a simple, feedforward ReLU-based multi-layer perception inherited from Neural Scene Flow Prior~\citep{nsfp}. How does this choice of learnable function class impact the performance of \ourmethod{}? To better understand these design choices we examine the choice of non-linearity and time feature encoding. 

\begin{figure}[htb]
    \centering
    \includegraphics{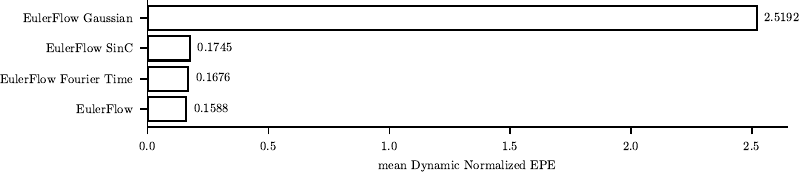}
    \caption{Mean Dynamic Normalized EPE of \ourmethod{} on the Argoverse 2 val split for less-smooth configurations of its learnable function class. These results indicate that the smoothness of the ReLU non-linearity proposed by \citeauthor{nsfp} transfers well to \ourmethod{}.
    }
    \figlabel{nonlinearitytime}
\end{figure}

One of \citeauthor{nsfp}'s core theoretical contributions demonstrates that NSFP's ReLU MLP is a good prior for scene flow because it represents a smooth learnable function class, and scene flow is often locally smooth with respect to input position. However, unlike NSFP, \ourmethod{} is fitting flow over a full ODE; while it seems reasonable to assume that this ODE is typically also locally smooth, cases like adjacent cars moving rapidly in opposite directions may benefit from the ability to model higher frequency, less locally smooth functions. To test this hypothesis, we ablate \ourmethod{} by replacing its normalized time with higher frequency sinusoidal time embeddings (mirroring \citeauthor{ntp}'s proposed time embedding for NTP), as well as try other popular non-linearities like SinC~\citep{ramasinghe2024on} and Gaussian~\citep{garf} from the coordinate network literature. \figref{nonlinearitytime} features negative results on these ablations across the board; Gaussians were unable to converge due the extremely high frequency representation triggering early stopping, while the use of SinC and higher frequency time embeddings both resulted in worse overall performance, indicating that \citeauthor{nsfp}'s smooth function prior does indeed seem appropriate for \ourmethod{}'s neural prior.

\subsection{\ourmethod{} with Monocular Depth Estimates}\appendixlabel{monodepth}

While \ourmethod{} only consumes point clouds, we can leverage RGB-based video monocular depth estimators to fit scene flow. In \figref{monodepth}, we use DepthCrafter~\citep{hu2024DepthCrafter} to generate a point cloud from the raw RGB of the tabletop video from \figref{orbbec}, Row 4.

\begin{figure}[htb]
    \centering

%     \begin{subfigure}[b]{0.24\textwidth}
%     \centering
%     \includegraphics[width=\textwidth]{screenshot_figures/waving_ball/screenshot_frame_0000.png}
% \end{subfigure}%
% \begin{subfigure}[b]{0.24\textwidth}
%     \centering
%     \includegraphics[width=\textwidth]{screenshot_figures/waving_ball/screenshot_frame_0013.png}
% \end{subfigure}%
% \begin{subfigure}[b]{0.24\textwidth}
%     \centering
%     \includegraphics[width=\textwidth]{screenshot_figures/waving_ball/screenshot_frame_0026.png}
% \end{subfigure}%
% \begin{subfigure}[b]{0.24\textwidth}
%     \centering
%     \includegraphics[width=\textwidth]{screenshot_figures/waving_ball/screenshot_frame_0039.png}
% \end{subfigure}

    \begin{subfigure}[b]{0.24\textwidth}
    \centering
    \includegraphics[width=\textwidth]{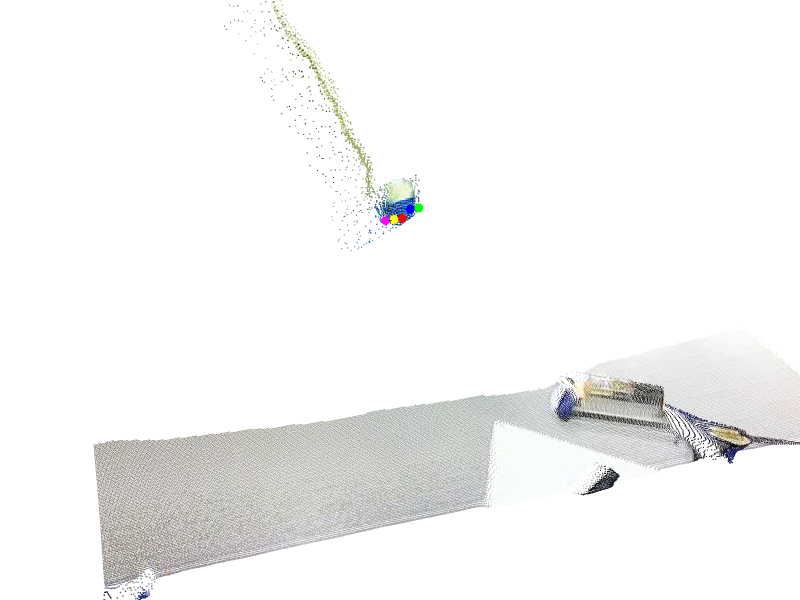}
\end{subfigure}%
\begin{subfigure}[b]{0.24\textwidth}
    \centering
    \includegraphics[width=\textwidth]{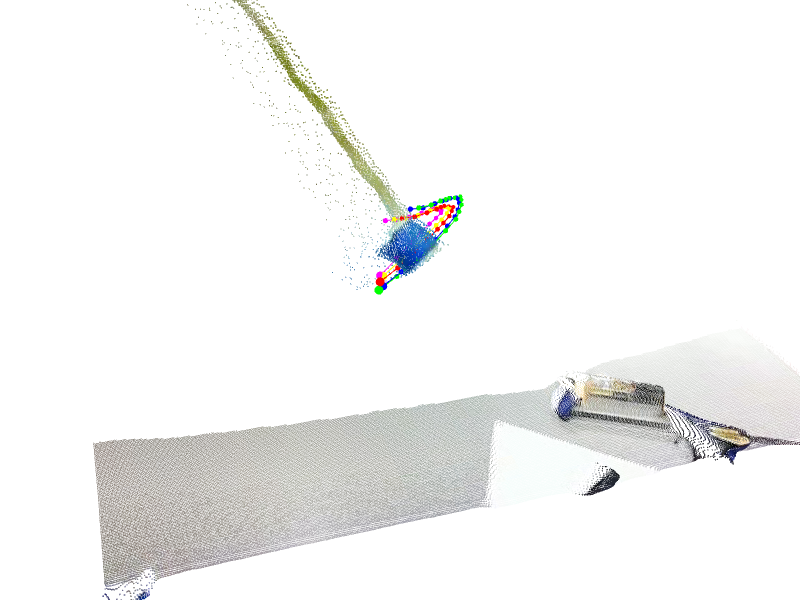}
\end{subfigure}%
\begin{subfigure}[b]{0.24\textwidth}
    \centering
    \includegraphics[width=\textwidth]{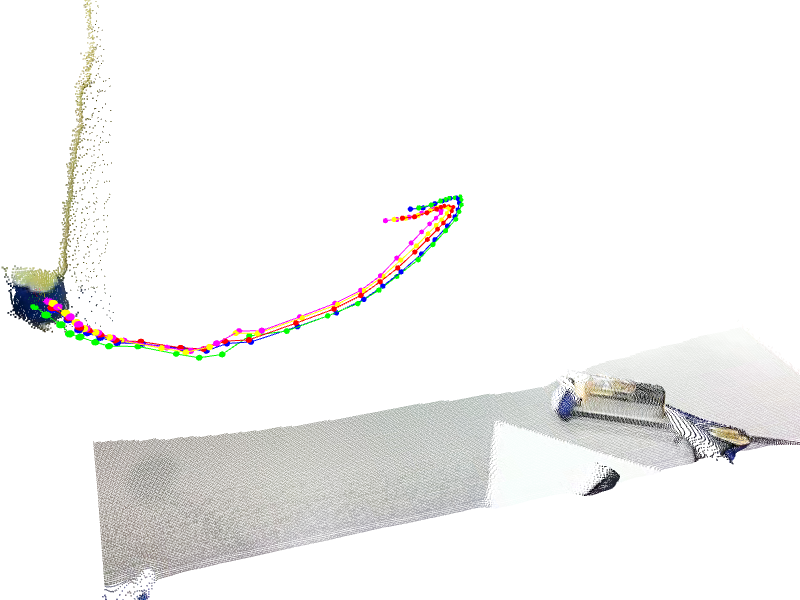}
\end{subfigure}%
\begin{subfigure}[b]{0.24\textwidth}
    \centering
    \includegraphics[width=\textwidth]{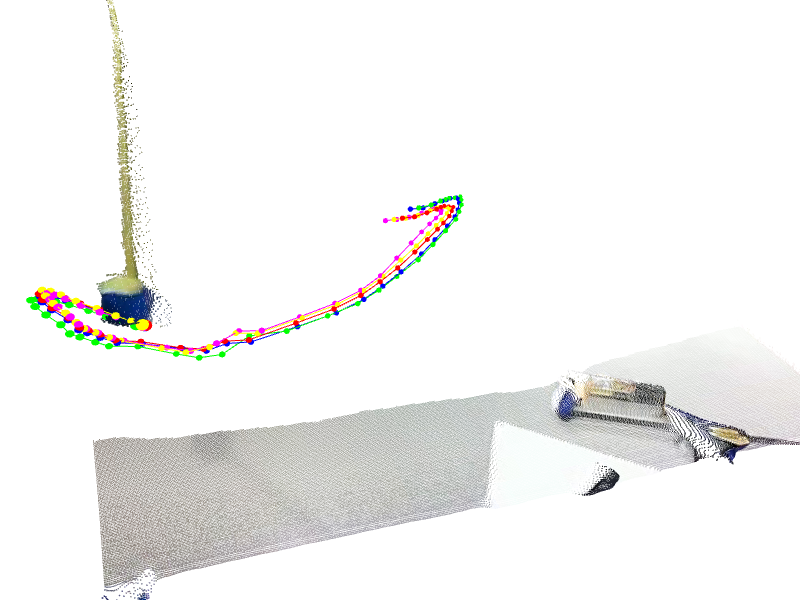}
\end{subfigure}

    \caption{Visualizations of \ourmethod{}'s emergent 3D point tracking behavior on monocular depth estimates from DepthCrafter~\citep{hu2024DepthCrafter}. Interactive visualizations available at \websitelink{}.}
    \figlabel{monodepth}
\end{figure}

%\neehar{Add Section on Low-Frame Rate EulerFlow}

\subsection{How Does \ourmethod{} Fail?}
\begin{figure}[htb]
    \centering

    \includegraphics[width=\linewidth, clip, trim={1.5cm 11cm 4cm 5cm}]{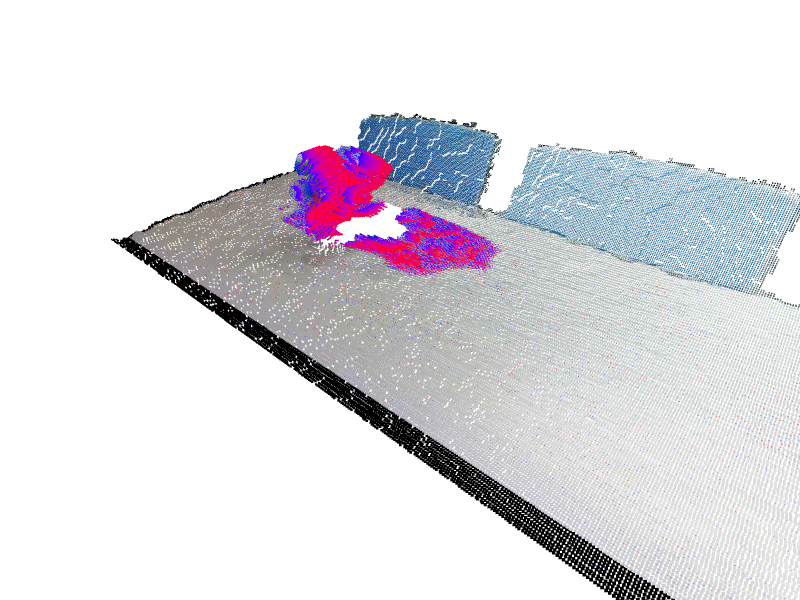}

    \caption{Visualizations of one of the failure modes of \ourmethod{} where flow is predicted on the edges of the moving "shadow" in the point cloud. Interactive visualizations available at \websitelink{}.}
    \figlabel{failures}
\end{figure}

As we discuss in \sectionref{limitations}, \ourmethod{} does not understand projective geometry --- its optimization losses use Chamfer Distance which directly associates points, sometimes resulting in moving shadows on background objects. To demonstrate this, we select a particularly egregious example in \figref{failures}, featuring a frame from the jack being thrown across the table. Due to the moving shadow cast by the jack onto the table, \ourmethod{} incorrectly assigns flow to the table surface nearby the jack, particularly on the leading edge, even though the table surface is stationary.

\section{FAQ}

\subsection{What datasets did you pretrain on?}

\ourmethod{} is not pretrained on any datasets. It is a test-time optimization method (akin to NeRFs), and as we show with our tabletop data, this means it runs out-of-the-box on arbitrary point cloud data.

\subsection{Why didn't you use a Neural ODE or a Liquid Neural Network?}

Neural ODEs~\citep{neuralode} take variable size and number of steps in latent space to do inference; imagine a ResNet that can use an ODE solver to dynamically scale the impact of the residual block, as well as decide the number of residual blocks. They are not a function class specially designed to fit derivative estimates well. Similar to Neural ODEs, Liquid Neural Networks~\citep{liquidneuralnet} focus on the same class of problems and are similarly not applicable.

\subsection{Why didn't you do experiments on FlyingThings3D / <simulated dataset>?}

Most popular synthetic datasets do not contain long observation sequences~\citep{flyingthings,Butler:ECCV:2012}, but instead include standalone frame pairs. Our method leverages the long sequence of observations to refine our neural estimate of the true ODE. Indeed, on two frames, \ourmethod{} collapses to NSFP.

More importantly, these datasets are also not representative of real world environments. To quote \citeauthor{chodosh2023}: ``[FlyingThings3D has] unrealistic rates of dynamic motion, unrealistic correspondences, and unrealistic sampling patterns. As a result, progress on these benchmarks is misleading and may cause researchers to focus on the wrong problems.'' \citeauthor{khatri2024trackflow} also make this point by highlighting the importance of meaningfully breaking down the object distribution during evaluation identify performance on rare safety-critical categories. FlyingThings3D does not have meaningful semantics; it's not obvious what things even matter or how to appropriately break down the scene.

Instead, we want to turn our attention to the sort of workloads that \emph{do} clearly matter --- describing motion in domains like manipulation or autonomous vehicles, where it seems clear that scene flow, if solved, will serve as powerful primitive for downstream systems. This is why we performed qualitative experiments on the tabletop data we collected ourselves; to our knowledge, no real-world dynamic datasets of this nature exist with ground truth annotations, but we want to emphasize that \ourmethod{} works in such domains, and consequently \ourmethod{} and other \ourpipelinefull{}-based methods can be used as a primitive in these real world domains.

\section{\ourmethod{} implementation details}\appendixlabel{implementationdetails}

Our neural prior $\network$ is a straightforward extension to NSFP's coordinate network prior\footnote{Hyperparameters (e.g.\ filter width of 128) of NSFP's prior are kept fixed, except for depth (\sectionref{mlpdepth}).}; however, instead of taking a 3D space vector (positions $X, Y, Z \in \mathbb{R}$) as input, we encode a 5D space-time-direction vector: positions $X,Y, Z, \in \mathbb{R}$, sequence normalized time $t \in [-1, 1]$ (i.e.\ the point cloud time scaled to this range), and direction $d \in \{\dirbackward = -1, \dirforward = 1\}$. This simple encoding scheme enables description of arbitrary regions of the ODE, allowing for the ODE to be queried at frequencies different from the sensor frame rate. Euler integration enables simple implementation of multi-step forward, backward, and cyclic consistency losses without extra bells and whistles. For efficiency, we use Euler integration with $\Delta{}t$ set as the time between observations for our ODE solver, enabling support for arbitrary sensor frame rates, and set the cycle consistency balancing term $\alpha = 0.01$ and optimization window $W=3$ for all experiments.

\section{\ourmethod{}'s ODE Derivation}\appendixlabel{derivation}

\subsection{Formulating the ODE}

Given a (possibly moving) particle in some canonical frame (i.e.\ time $0$), we define a function $L(x_0, y_0, z_0, t)$ that can describe its location at an arbitrary future time $t$, i.e.\ a Lagrangian description of motion (\figref{euler_vs_lagrange}).

\begin{equation}
  L(x_0, y_0, z_0, t ) = x_t, y_t, z_t  
\end{equation}

For notational clarity to access $x_t, y_t, z_t$ individually, we can define 

\begin{align}
  L_x(x_0, y_0, z_0, t) &= x_t\\  
  L_y(x_0, y_0, z_0, t) &= y_t\\
  L_z(x_0, y_0, z_0, t) &= z_t
\end{align}

Similarly, we can define $F(x_t, y_t, z_t, t)$ to describe the instantaneous velocity of a point $x_t,y_t,z_t$ at some arbitrary time $t$, i.e.\ a Eulerian description of motion (\figref{euler_vs_lagrange}).

\begin{equation}
  \frac{d L(x_0, y_0, z_0, t)}{d t} = \frac{d L}{d t} = \left(\frac{d L_x}{d t}, \frac{d L_y}{d t},\frac{d L_z}{d t}\right) = F(x_t , y_t, z_t, t)  
\end{equation}

$F$ is defined in terms of the total derivative of $L$ with respect to $t$, as $x_0, y_0, z_0$ are initial conditions that do not vary with time (i.e. $\frac{dL}{dt} = \frac{\partial L}{\partial t} + \frac{\partial L}{\partial x_0} \frac{d x_0}{d t} + \frac{\partial L}{\partial y_0} \frac{d y_0}{d t} + \frac{\partial L}{\partial z_0} \frac{d z_0}{d t} = \frac{\partial L}{\partial t}$, as $\frac{dx_0}{dt} = \frac{dy_0}{dt} = \frac{dz_0}{dt} = 0$). We can exactly define $L$ recursively in terms of the initial conditions and $F$, i.e.

\begin{equation}
  L(x_0, y_0, z_0, t) = \left({x_0, y_0, z_0}\right) + \int_0^t F(L_x(x_0, y_0, z_0, \tau), L_y(x_0, y_0, z_0, \tau), L_z(x_0, y_0, z_0, \tau), \tau)d\tau   
\end{equation}

or, more compactly,

\begin{equation}
  L(x_0, y_0, z_0, t) = \left({x_0, y_0, z_0}\right) + \int_0^t F(x_\tau, y_\tau, z_\tau, \tau)d\tau
  \equationlabel{ldefinitionzerotot}
\end{equation}

Our function $L$ can thus be defined as a multi-dimensional ODE in terms of $F$ with initial conditions $x_0, y_0, z_0$. 

\subsection{Arbitrary start and end times from the Eulerian formulation}\appendixlabel{arbitrarystartandend}

In the above derivation, $L$ requires that a moving point be defined in terms of a canonical frame defined at time $0$, as is common in the deformation in reconstruction literature. However, the Eulerian formulation has no such requirement, allowing us to select arbitrary start and end times across different point queries. To showcase this, we can query $F$ to extract the trajectory of a particle at $t$ across the range $[t,t']$ starting at $x_t, y_t, z_t$ simply by changing the range of the integral in \equationref{ldefinitionzerotot}, i.e.

\begin{equation}
 E(x_t, y_t, z_t, t, t') = \left({x_t, y_t, z_t}\right) + \int_{t}^{t'} F(x_\tau, y_\tau, z_\tau, \tau)d\tau   
  \equationlabel{rangeshifttrick}
\end{equation}

While $E$ and $L$ appear similar on their face, $E$ is strictly more flexible than $L$. In principle you could choose to redefine $L$ to use $t$ as the time for your canonical frame, but this is a \emph{global} choice; you cannot do this on a per-query basis. However, with $E$'s  Eulerian framing, we can extract a different point's trajectory from the entirely different range $t^\dagger$ to $t^\ddagger$ (i.e.\ $E(x_{t^\dagger}, y_{t^\dagger}, z_{t^\dagger}, t^\dagger, t^\ddagger)$) without concern for a canonical frame definition. It need not even be the case that $t < t'$; indeed, this extraction works even if $t > t'$, i.e.\ extracting the backwards trajectory through time.

\subsection{Euler Integration to approximately solve the ODE}

If $F$ is of arbitrary form and we want to compute the concrete values of $L$, we cannot exactly compute the continuous integral from $0$ to $t$; we must approximate this with finite differences. Thus, we split the time range $0$ to $t$ into $k$ steps, where each step is of size $\frac{t}{k}$. Thus, we can again define $L$ via recursion, but this time explicitly.

\begin{align}
  L(x_0, y_0, z_0, 0) &= \left({x_0, y_0, z_0}\right) \\
  L(x_0, y_0, z_0, \tau + \frac{t}{k}) &\approx L(x_0, y_0, z_0, \tau) + \frac{t}{k} \cdot F(x_{\tau}, y_{\tau}, z_{\tau}, \tau), 
\end{align}

or directly without recursion,

\begin{equation}
    L(x_0, y_0, z_0, t) \approx \left({x_0, y_0, z_0}\right) + \sum_{n=1}^k \frac{t}{k} \cdot F(x_{n \frac{t}{k}}, y_{n \frac{t}{k}}, z_{n \frac{t}{k}}, n \frac{t}{k})
    \equationlabel{expliciteuler}
\end{equation}

This finite difference solving approach is Euler integration.

\subsection{Estimating the flow field with \ourmethod{}'s neural prior}\appendixlabel{neuralpriorrollout}

For a given scene, we do not have access to $L$ or $F$ directly; these are are the \emph{true} functions that uniquely characterize the underlying motion of the scene that we are trying to estimate. For \ourmethod{}, we represent our estimate of the scene's flow field $F$ with a neural prior, $\network$, i.e. 

\begin{equation}
F(x,y,z,t) \approx \network(x,y,z,t)
\equationlabel{eulerflownetworkequiv}
\end{equation}

and thus

\begin{equation}
    L(x_0, y_0, z_0, t) \approx \left({x_0, y_0, z_0}\right) + \sum_{n=1}^k \frac{t}{k} \cdot \network{}(x_{n \frac{t}{k}}, y_{n \frac{t}{k}}, z_{n \frac{t}{k}}, n \frac{t}{k})
    \equationlabel{expliciteulernetworkequiv}
\end{equation}

and, using the arbitrary start and end definition from \appendixref{arbitrarystartandend}, with $k$ steps from the range $t$ to $t'$ and $\delta = \frac{t' - t}{k}$

\begin{equation}
    E(x_t, y_t, z_t, t, t') \approx E_{\network}(x_t, y_t, z_t, t, t') = \left({x_t, y_t, z_t}\right) + \sum_{n=1}^{k} \delta \cdot \network{}(x_{n \delta + t}, y_{n \delta + t}, z_{n \delta + t}, n \delta + t)
    \equationlabel{eulertrajectory}
\end{equation}

This formulation makes \ourmethod{} highly flexible, enabling optimization of $\network$'s estimate of $F$ with objectives that take either an Eulerian view (directly on $\network$ via \equationref{eulerflownetworkequiv}) or a Lagrangian view (on point rollouts for arbitrary start and end ranges via \equationref{eulertrajectory}).

\end{document}

%% file: math_commands.tex
\usepackage{amsmath,amsfonts,bm}

\def\eqref#1{equation~\ref{#1}}
\def\1{\bm{1}}

\DeclareMathAlphabet{\mathsfit}{\encodingdefault}{\sfdefault}{m}{sl}
\SetMathAlphabet{\mathsfit}{bold}{\encodingdefault}{\sfdefault}{bx}{n}

%% file: abstract_hardstyle.tex
%We present a step function unlock in unsupervised scene flow capabilities. 
We reframe scene flow as the task of estimating a continuous space-time ordinary differential equation (ODE) that describes motion for an entire observation sequence, represented with a neural prior. Our method, \emph{\ourmethod{}}, optimizes this neural prior estimate against several multi-observation reconstruction objectives, enabling high quality scene flow estimation via self-supervision on real-world data. \ourmethod{} works out-of-the-box without tuning across multiple domains, including large-scale autonomous driving scenes and dynamic tabletop settings. Remarkably, \ourmethod{} produces high quality flow estimates on small, fast moving objects like birds and tennis balls, and exhibits emergent 3D point tracking behavior by solving its estimated ODE over long-time horizons. On the Argoverse 2 2024 Scene Flow Challenge, \ourmethod{} outperforms \emph{all} prior art, surpassing the next-best \emph{unsupervised} method by more than $2.5\times$, and even exceeding the next-best \emph{supervised} method by over 10\%. See \websitelink{} for interactive visuals.

%% file: tex_figures/two_frame_vs_full_sequence_2x4.tex
\newcommand{\seqlencaptionfontsize}{\fontsize{7}{7}\selectfont}

\begin{figure}[t]
\centering
% First row: Two Frame figures
\begin{subfigure}[b]{0.24\textwidth}
    \centering
    \includegraphics[width=\textwidth]{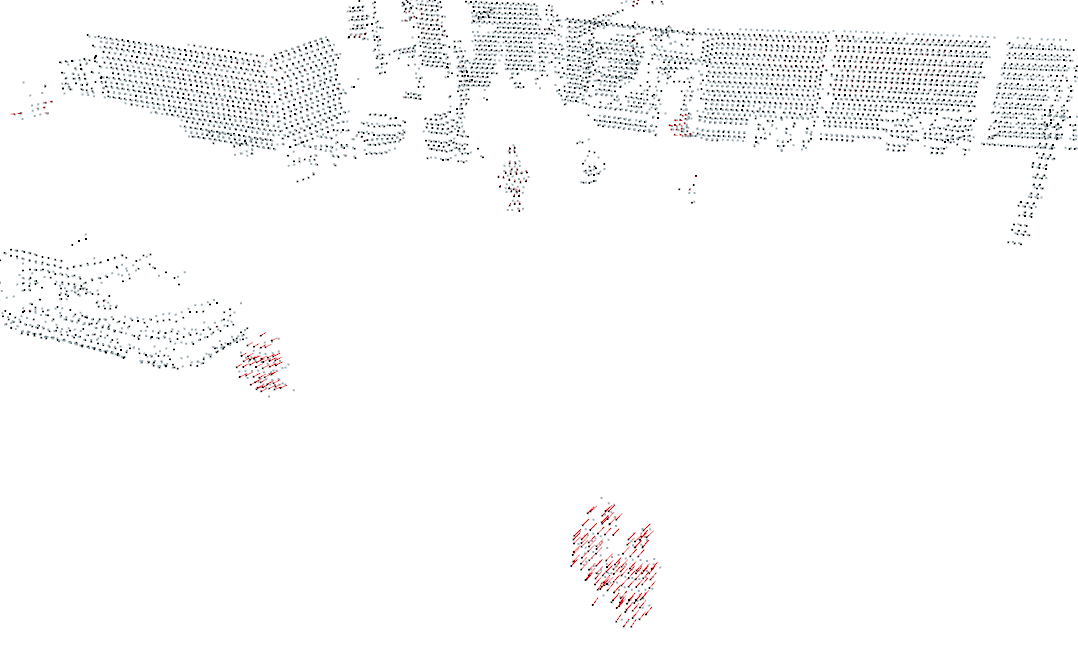}
    \caption{\seqlencaptionfontsize \ourmethod{} (Two Frame)}
    \figlabel{flyingbirdgigachad_twoframe}
\end{subfigure}%
\begin{subfigure}[b]{0.24\textwidth}
    \centering
    \includegraphics[width=\textwidth]{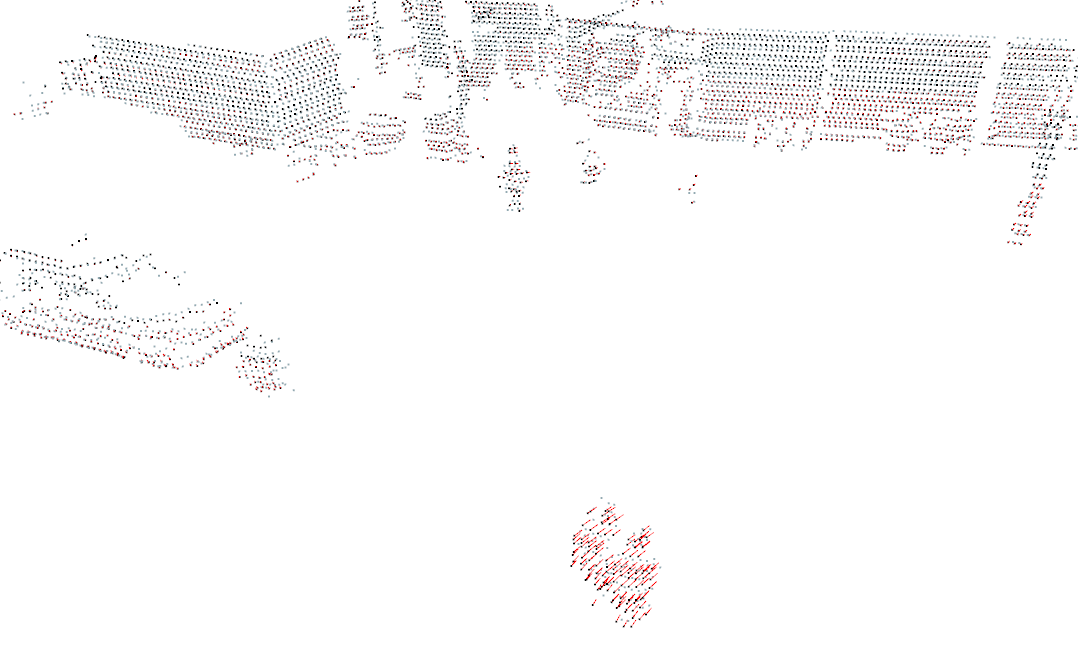}
    \caption{\seqlencaptionfontsize Fast NSF (Two Frame)}
    \figlabel{flyingbirdfastnsf_twoframe}
\end{subfigure}%
\begin{subfigure}[b]{0.24\textwidth}
    \centering
    \includegraphics[width=\textwidth]{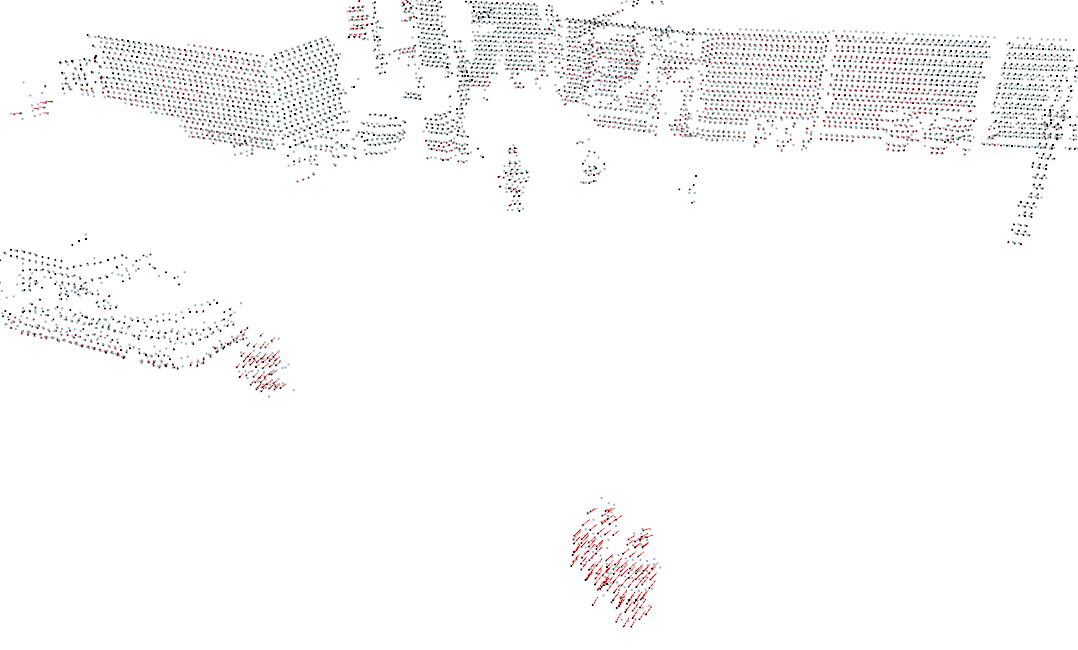}
    \caption{\seqlencaptionfontsize \citeauthor{liu2024selfsupervisedmultiframeneuralscene} (Two Frame)}
    \figlabel{flyingbirdliuetal_twoframe}
\end{subfigure}%
\begin{subfigure}[b]{0.24\textwidth}
    \centering
    \includegraphics[width=\textwidth]{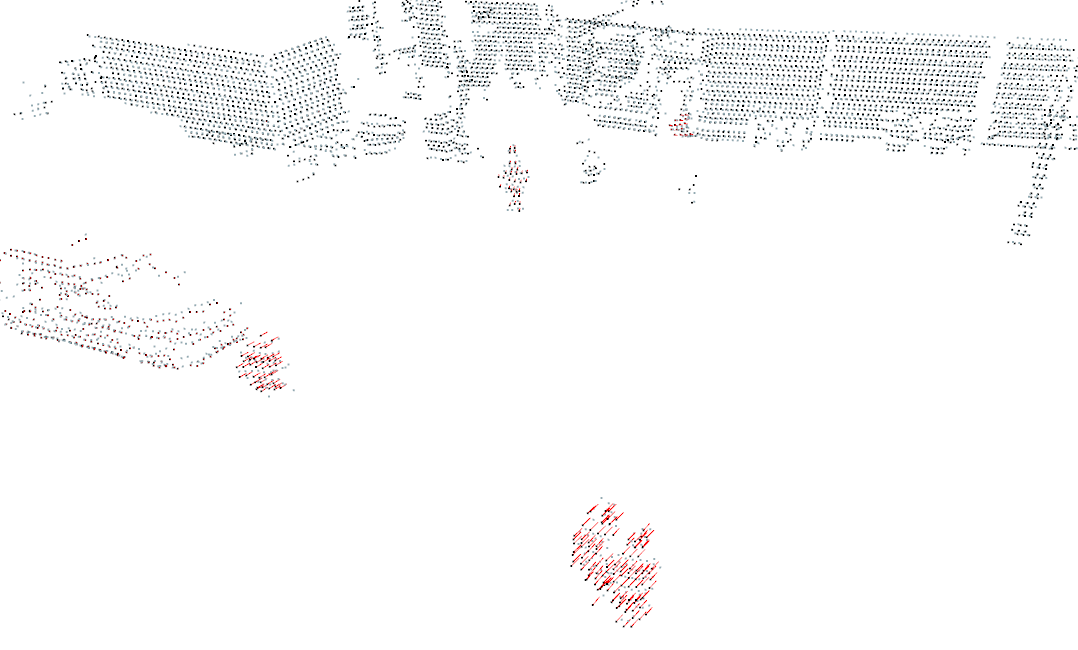}
    \caption{\seqlencaptionfontsize Ground Truth (Two Frame)}
    \figlabel{flyingbirdgroundtruth_twoframe}
\end{subfigure}

% Second row: Full Sequence figures
\begin{subfigure}[b]{0.24\textwidth}
    \centering
    \includegraphics[width=\textwidth]{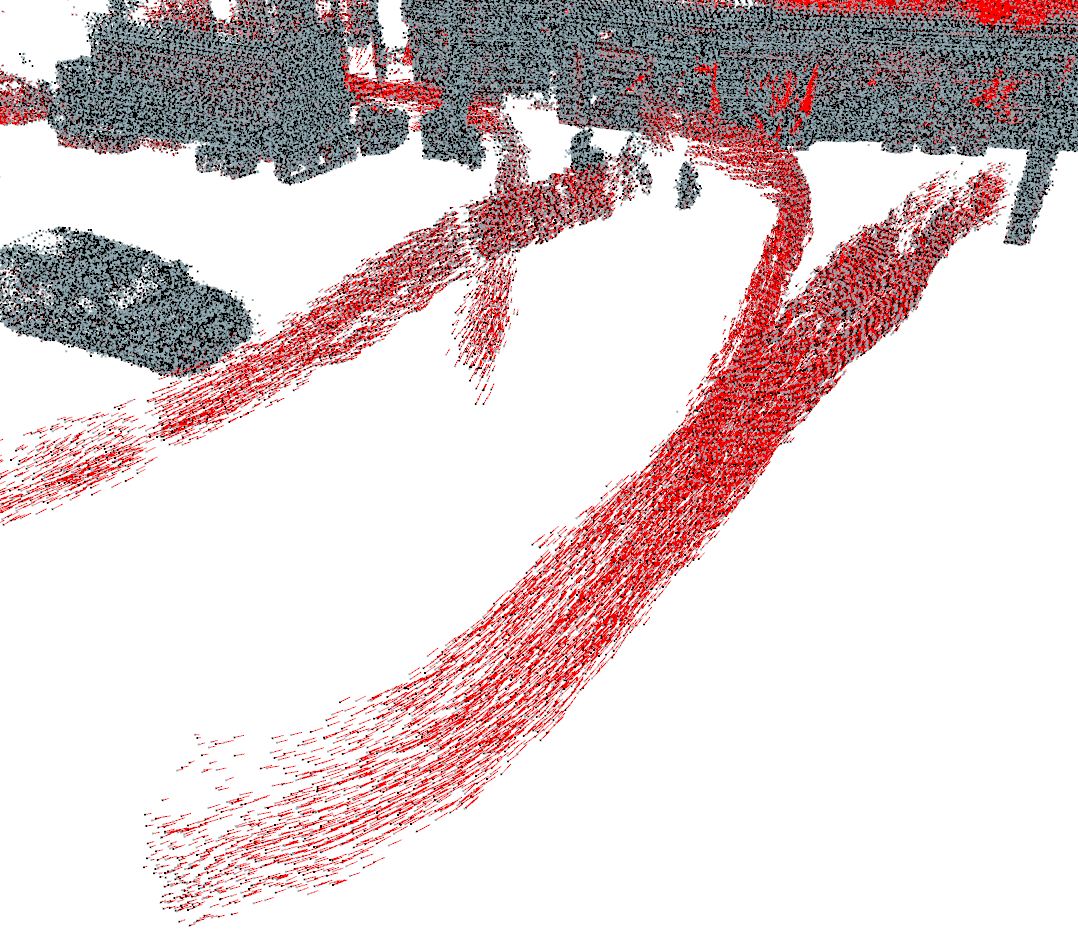}
    \caption{\seqlencaptionfontsize \ourmethod{} (Full Sequence)}
    \figlabel{flyingbirdgigachad_fullsequence}
\end{subfigure}%
\begin{subfigure}[b]{0.24\textwidth}
    \centering
    \includegraphics[width=\textwidth]{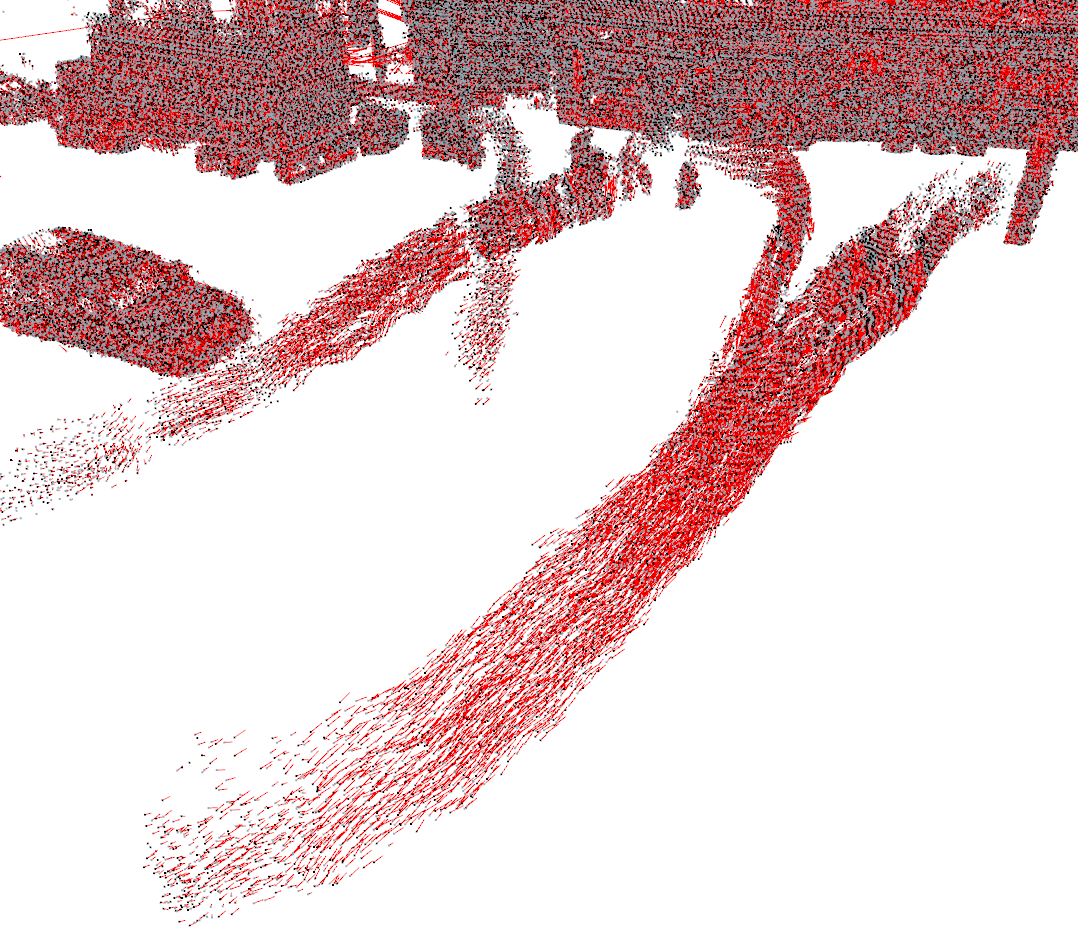}
    \caption{\seqlencaptionfontsize Fast NSF (Full Sequence)}
    \figlabel{flyingbirdfastnsf_fullsequence}
\end{subfigure}%
\begin{subfigure}[b]{0.24\textwidth}
    \centering
    \includegraphics[width=\textwidth]{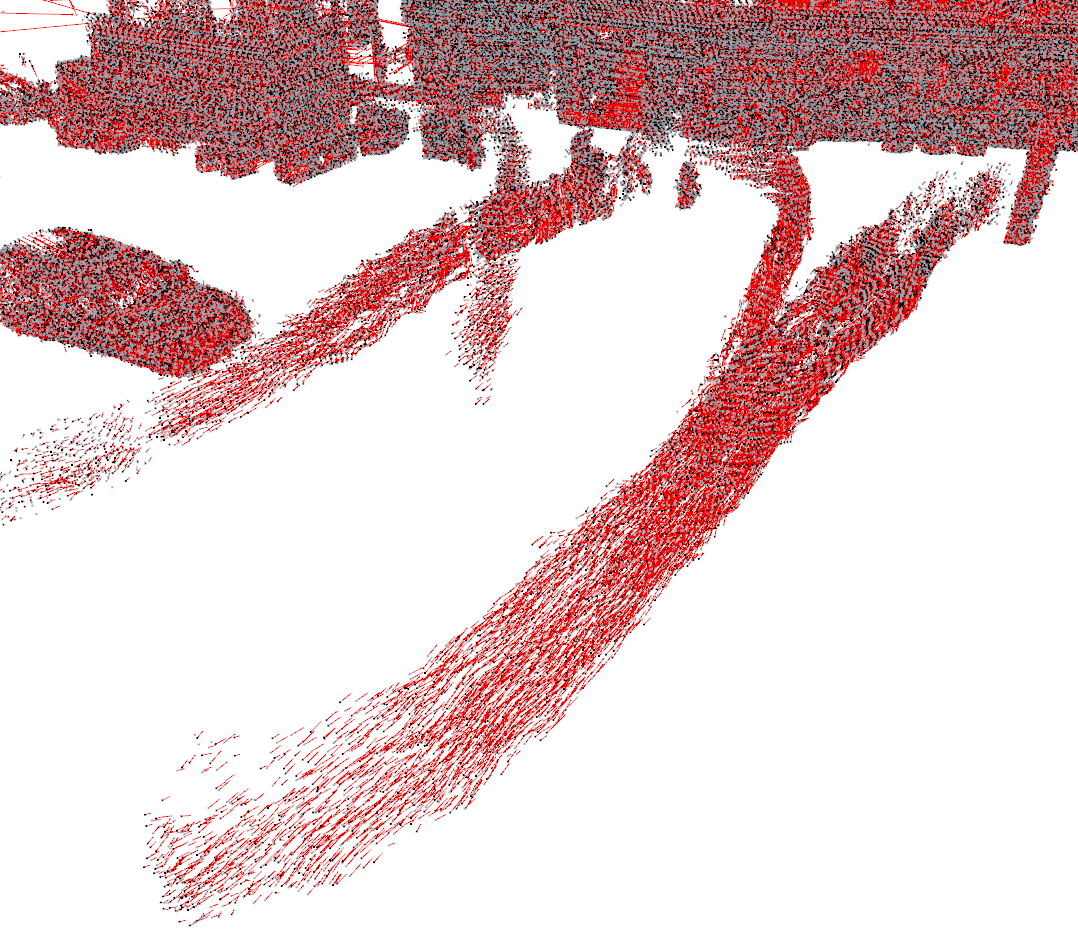}
    \caption{\seqlencaptionfontsize \citeauthor{liu2024selfsupervisedmultiframeneuralscene} (Full Sequence)}
    \figlabel{flyingbirdliuetal_fullsequence}
\end{subfigure}%
\begin{subfigure}[b]{0.24\textwidth}
    \centering
    \includegraphics[width=\textwidth]{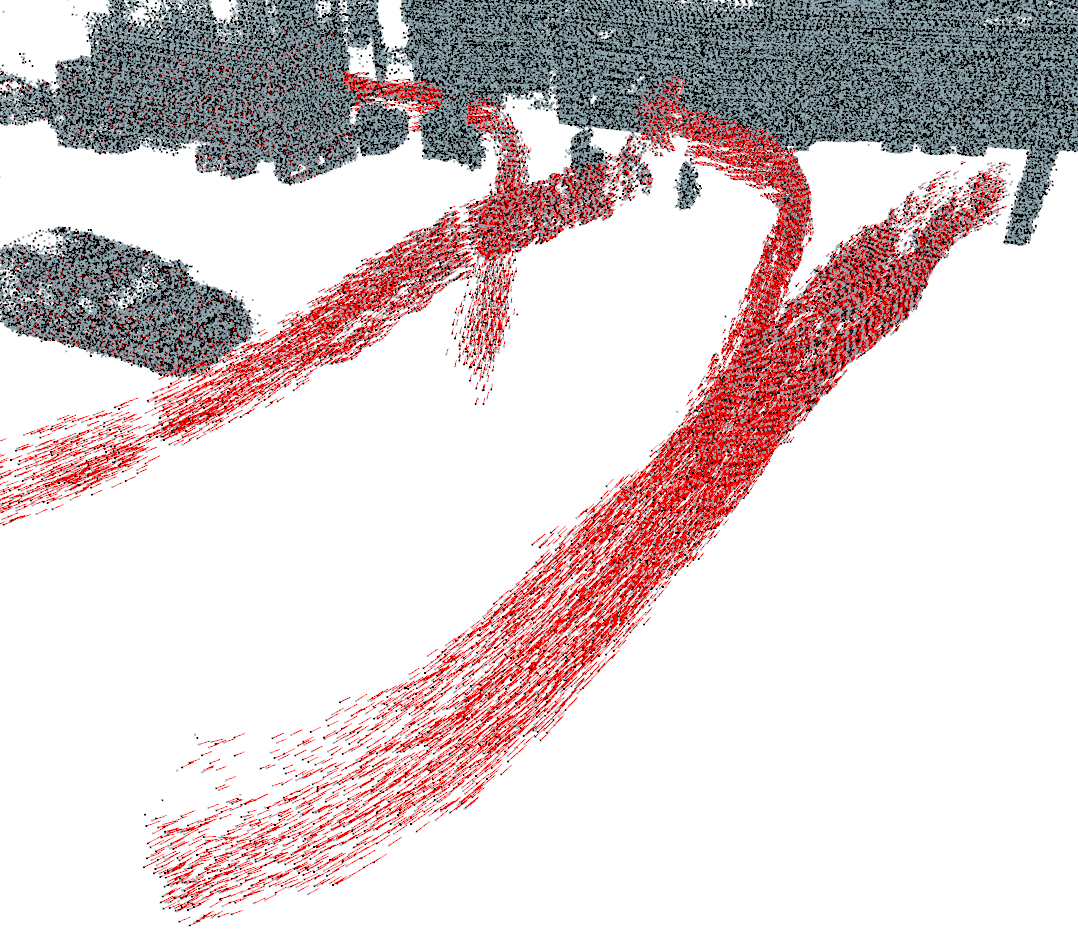}
    \caption{\seqlencaptionfontsize Ground Truth (Full Sequence)}
    \figlabel{flyingbirdgroundtruth_fullsequence}
\end{subfigure}

\caption{ We visualize an example of five pedestrians crossing the street in front of a stopped car, cherrypicked to have unusually high density lidar returns, making it particularly easy to estimate flow. \figrefs{flyingbirdgigachad_twoframe}{flyingbirdgroundtruth_twoframe} depict a two-frame flow visualization of \ourmethod{} and several strong baselines. Notably, only visualizing flow over two frames makes it difficult to distinguish flow quality. In contrast,  \figrefs{flyingbirdgigachad_fullsequence}{flyingbirdgroundtruth_fullsequence} depict flow vectors over the full sequence, making differences in quality clear; for example, \ourmethod{} is the only one without artifacts on the stopped car.\vspace{-2em}}
\figlabel{walkingpeds}
\end{figure}

%% file: tex_figures/field_descriptions.tex
\begin{figure}[ht]
    \centering
    \begin{tikzpicture}[scale=0.5]

        % Eulerian section
        \node at (-3.5, 4.5) {\textbf{Eulerian View}};
        
        % Eulerian grid
        \draw[step=1cm,gray,very thin] (-5,1) grid (-2,4);
        
        % Arrows representing flow at grid points
        \draw[->,thick] (-4.5,3.5) -- (-4.5+0.5,3.5+0.3);
        \draw[->,thick] (-3.5,3.5) -- (-3.5+0.2,3.5+0.6);
        \draw[->,thick] (-2.5,3.5) -- (-2.5+0.3,3.5+0.2);
        
        \draw[->,thick] (-4.5,2.5) -- (-4.5+0.4,2.5+0.4);
        \draw[->,thick] (-3.5,2.5) -- (-3.5+0.6,2.5+0.1);
        \draw[->,thick] (-2.5,2.5) -- (-2.5+0.3,2.5+0.5);
        
        \draw[->,thick] (-4.5,1.5) -- (-4.5+0.2,1.5+0.4);
        \draw[->,thick] (-3.5,1.5) -- (-3.5+0.5,1.5+0.3);
        \draw[->,thick] (-2.5,1.5) -- (-2.5+0.4,1.5+0.4);
        
        % Lagrangian section
        \node at (4, 4.5) {\textbf{Lagrangian View}};
        
        % Pathlines for particles
        \draw[->,thick,blue] (2,1) .. controls (2.5,2) and (3,3) .. (4,4);
        \draw[->,thick,green] (2.5,1) .. controls (3,2) and (4,3) .. (5,4);
        \draw[->,thick,red] (3,1) .. controls (3.5,2) and (4.5,3) .. (6,4);

        % Particles
        \filldraw[blue] (2,1) circle (1pt) node[anchor=south] {\tiny$A_0$};
        \filldraw[green] (2.5,1) circle (1pt) node[anchor=south] {\tiny$B_0$};
        \filldraw[red] (3,1) circle (1pt) node[anchor=south] {\tiny$C_0$};

        % Particle positions at t2
        \filldraw[blue] (4,4) circle (1pt) node[anchor=north] {\tiny$A_t$};
        \filldraw[green] (5,4) circle (1pt) node[anchor=north] {\tiny$B_t$};
        \filldraw[red] (6,4) circle (1pt) node[anchor=north] {\tiny$C_t$};

    \end{tikzpicture}
    \caption{Comparison of Eulerian and Lagrangian descriptions of 2D flow. An Eulerian view characterizes a flow field via instantaneous velocities at many different points, while a Lagrangian view characterizes a flow field via trajectories of many different particles across time. Both approaches are valid ways of describing an underlying flow field, and with sufficient characterization one view can be readily converted to another, but the Lagrangian view relies on a definition of the definition of consistent canonical frame. \vspace{-1em}}
    \figlabel{euler_vs_lagrange}
\end{figure}

%% file: tex_figures/bird_two_rows.tex
\begin{figure}[htb]
\centering
% First row
\begin{subfigure}[b]{0.32\textwidth}
    \centering
    \includegraphics[width=\textwidth]{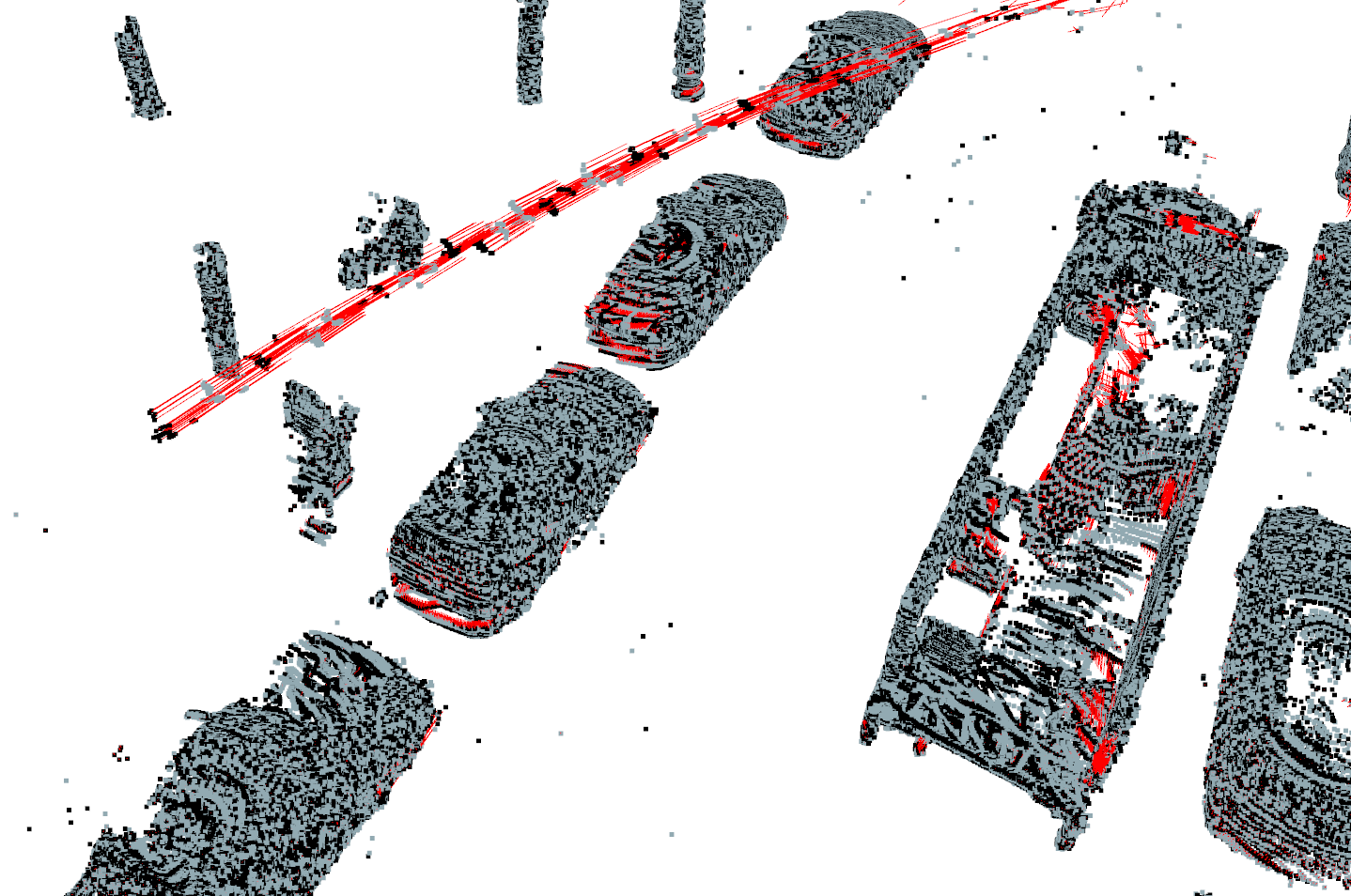}
    \caption{\ourmethod{} (Ours)}
    \figlabel{flyingbirdgigachad}
\end{subfigure}%
\begin{subfigure}[b]{0.32\textwidth}
    \centering
    \includegraphics[width=\textwidth]{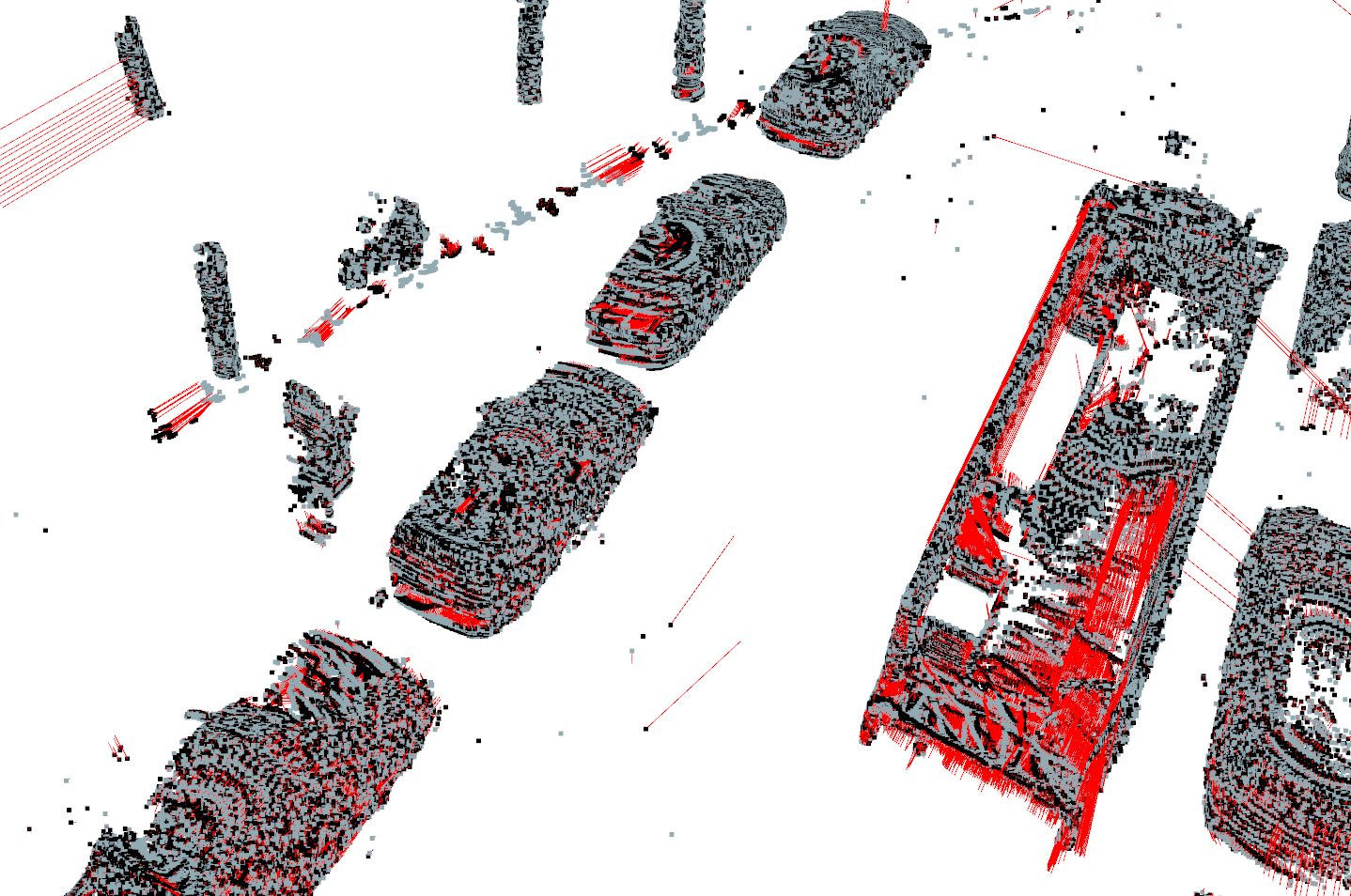}
    \caption{Fast NSF}
    \figlabel{flyingbirdfastnsf}
\end{subfigure}%
\begin{subfigure}[b]{0.32\textwidth}
    \centering
    \includegraphics[width=\textwidth]{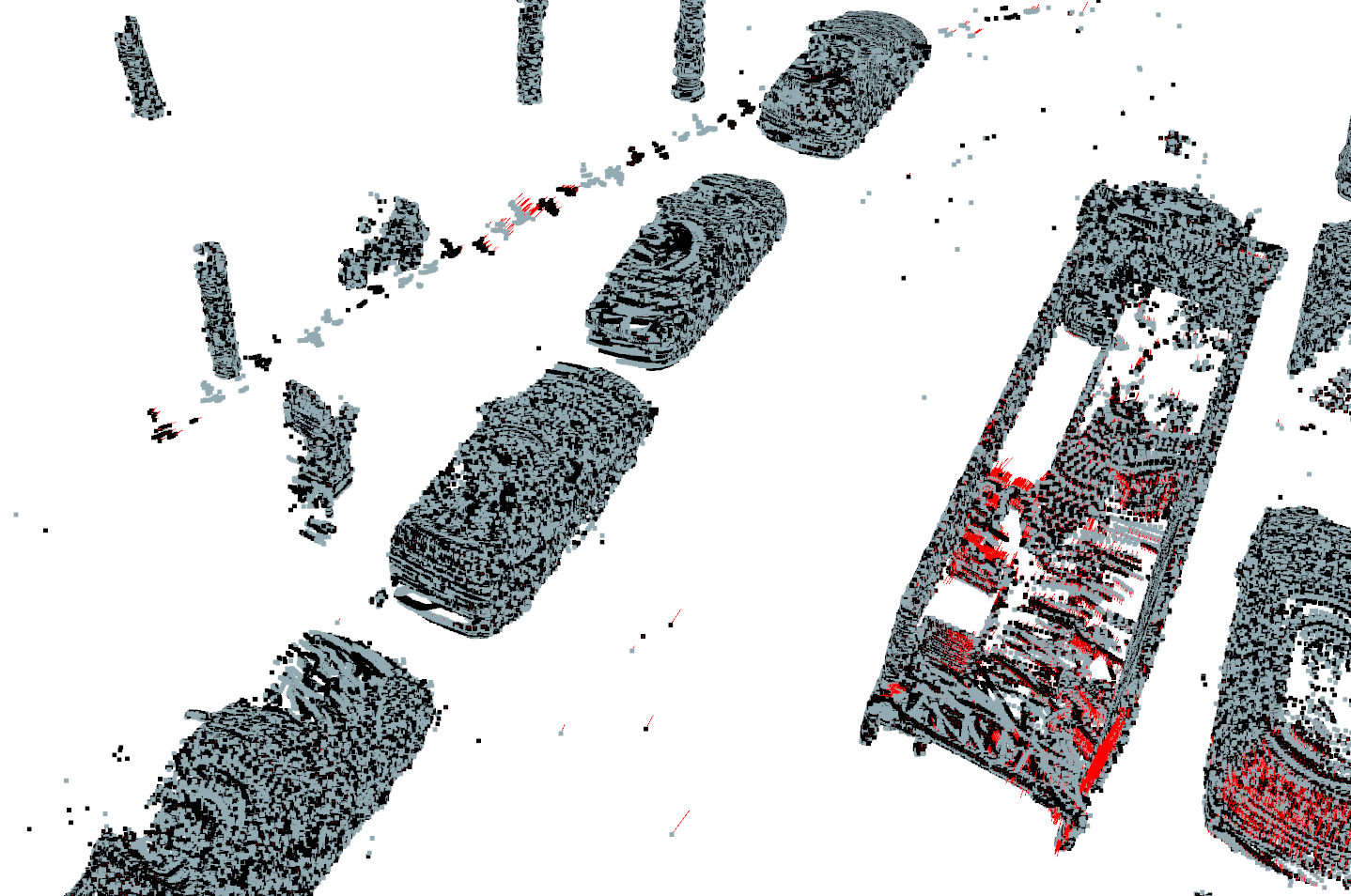}
    \caption{ZeroFlow 5x}
    \figlabel{flyingbirdzeroflow}
\end{subfigure}
% Second row
\begin{subfigure}[b]{0.32\textwidth}
    \centering
    \includegraphics[width=\textwidth]{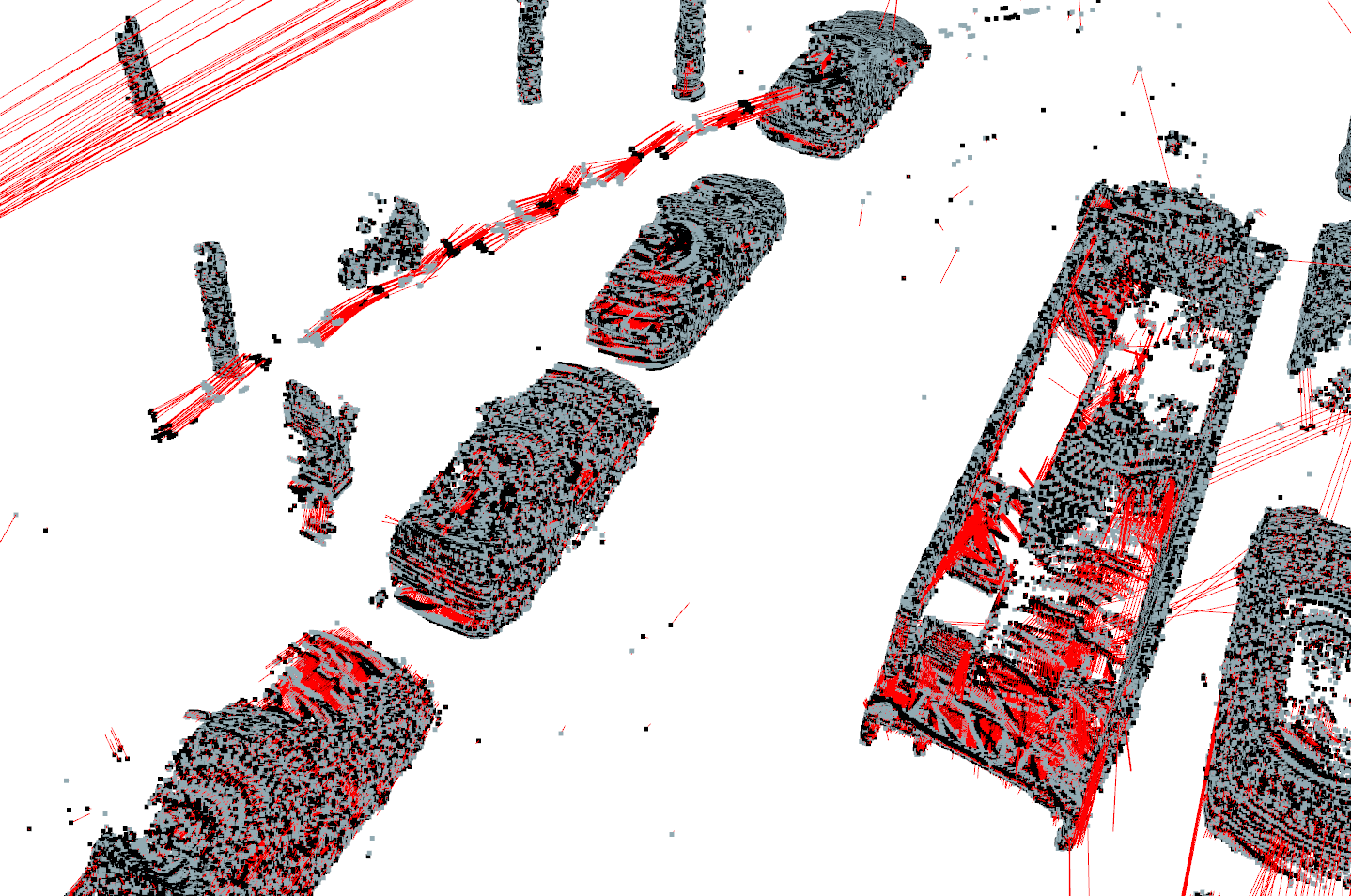}
    \caption{\citeauthor{liu2024selfsupervisedmultiframeneuralscene}}
    \figlabel{flyingbirdliuetal}
\end{subfigure}%
\begin{subfigure}[b]{0.32\textwidth}
    \centering
    \includegraphics[width=\textwidth]{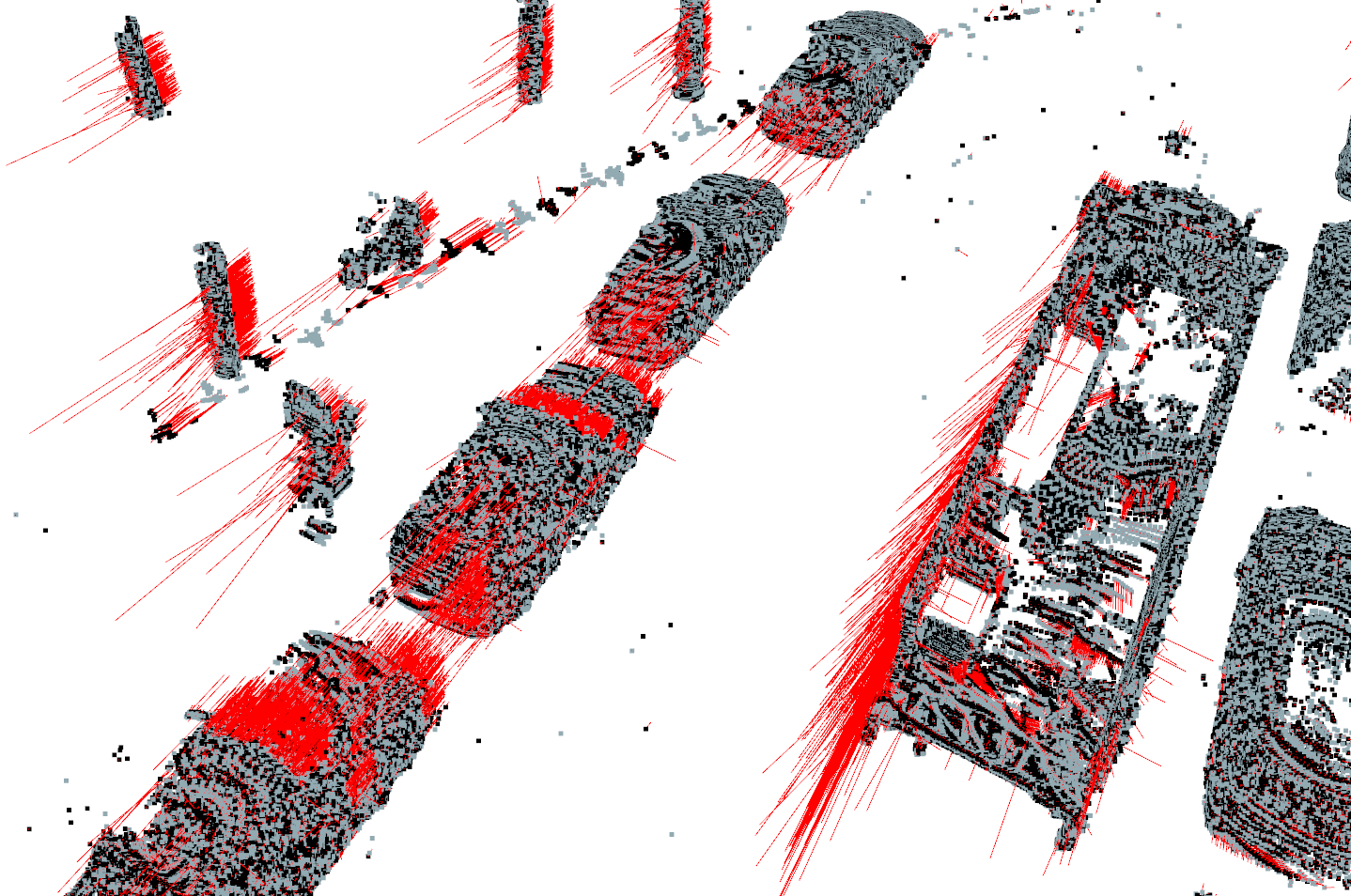}
    \caption{NSFP}
    \figlabel{flyingbirdnsfp}
\end{subfigure}%
\begin{subfigure}[b]{0.32\textwidth}
    \centering
    \includegraphics[width=\textwidth]{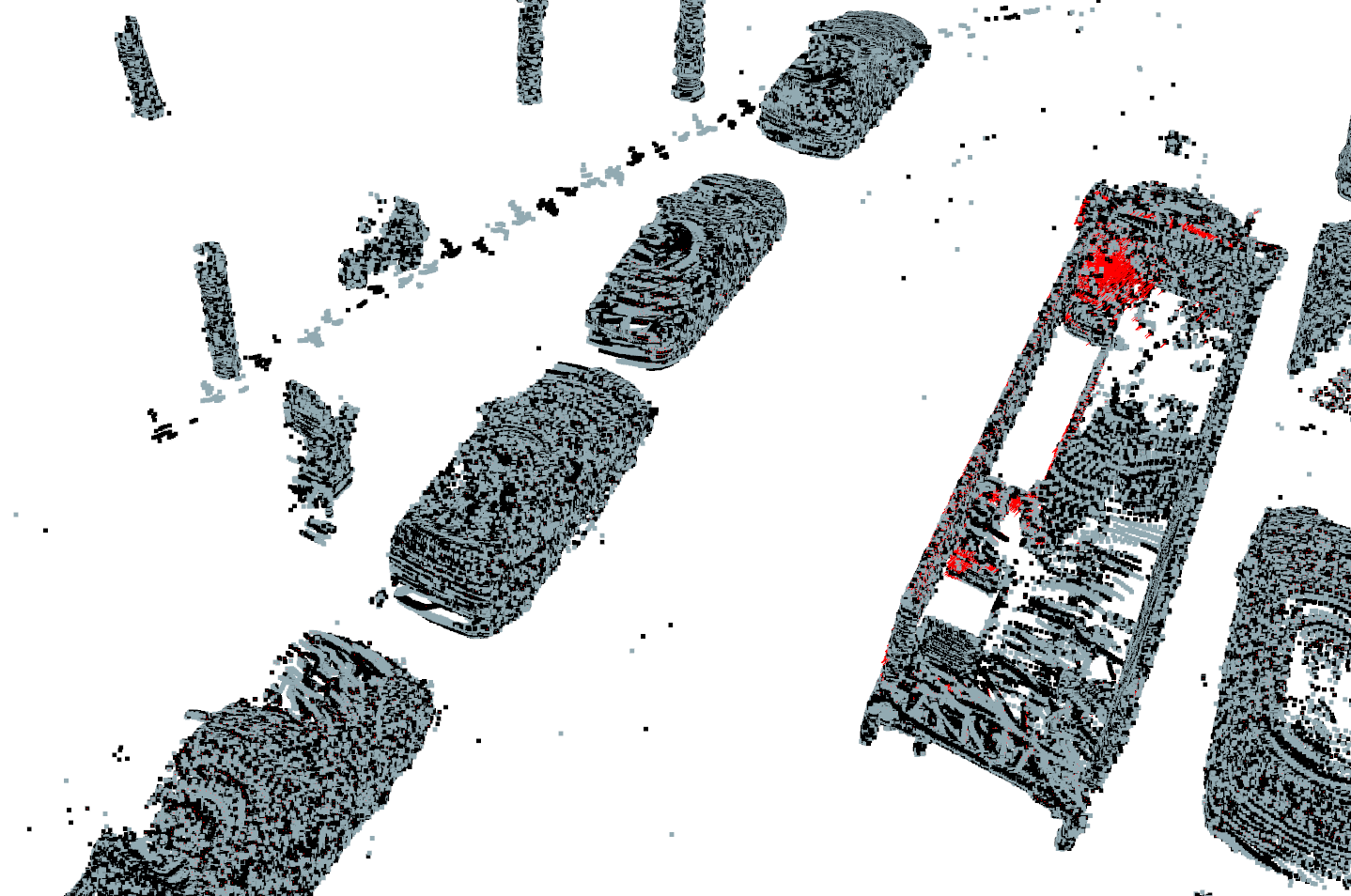}
    \caption{Ground Truth}
    \figlabel{flyingbirdgt}
\end{subfigure}
%\texttt{5f016e44-0f38-3837-9111-58ec18d1a5e6}
\caption{Visualization of \ourmethod{} compared to prior art for the same scene as \figref{teasersceneflow} and \figref{birdtrajectory}. \ourmethod{} is able to extract the bird's trajectory; however, all other methods except \citeauthor{liu2024selfsupervisedmultiframeneuralscene} fail to recognize this motion, and \citeauthor{liu2024selfsupervisedmultiframeneuralscene}'s flow is marred by severe scene artifacts. The bird is outside the labeled object taxonomy, and so its motion is unlabeled in the ground truth (\figref{flyingbirdgt}).\vspace{-1em}}
\figlabel{flyingbird}
\end{figure}

%% file: merged_references.bib
@article{scalablesceneflow,
  title     = {{Representation Learning for Object Detection from Unlabeled Point Cloud Sequences}},
    year={2022}
YEAR = {2018},

@article{seflow,
journal={arXiv preprint arXiv:1908.09492},
      title={GPT-4 Technical Report}, 
    journal   = {International Conference on Intelligent Robots and Systems (IROS)},
  booktitle={NeurIPS Datasets Track},
  author={Dong, Guanting and Zhang, Yueyi and Li, Hanlin and Sun, Xiaoyan and Xiong, Zhiwei},
title = {{Just Go With the Flow: Self-Supervised Scene Flow Estimation}},

@String(IROS = {Int. Conf. Intel. Rob. Sys.})

@String(CVPR= {IEEE Conf. Comput. Vis. Pattern Recog.})

@INPROCEEDINGS{9856954,
  journal      = {CoRR},

@String(ECCV= {Eur. Conf. Comput. Vis.})

@String(RAL= {IEEE Rob. Aut. Letters})

@article{nsfp,
      eprint={2303.11420},
and Dong, Shaocong
              Florian Golemo and Charles Herrmann and Thomas Kipf and Abhijit Kundu and

@inproceedings{flyingthings,
                  Piotr Doll{\'{a}}r and
  pages={1137-1149},}

@inproceedings{gu2019hplflownet,
    booktitle   = {Twelfth International Conference on Learning Representations (ICLR)},
  title={{Robust estimation of a location parameter}},

@inproceedings{battrawy2022rms,

@String(PAMI = {IEEE Trans. Pattern Anal. Mach. Intell.})

@article{Zhai2020FlowMOT3M,
  title={{Language Models are Unsupervised Multitask Learners}},
  author={Zermas, Dimitris and Izzat, Izzat and Papanikolopoulos, Nikolaos},
  journal = {arXiv preprint},

@String(ICLR = {Int. Conf. Learn. Represent.})

@inproceedings{tishchenko2020self,
  pages={132914--132927},

@inproceedings{wu2020pointpwc,
  title={{A solution for the best rotation to relate two sets of vectors}},
  author={Kingma, Diederik P and Ba, Jimmy},
 author = {Krizhevsky, Alex and Sutskever, Ilya and Hinton, Geoffrey E},
  volume={32},
  series    = {Proceedings of Machine Learning Research}
author = {Liu, Fangchen and Liu, Hao and Grover, Aditya and Abbeel, Pieter},
VOLUME = {18},
  author={Li, Hao and Sumner, Robert W and Pauly, Mark},
    booktitle = {ProceedTarashaings of the IEEE International Conference on Computer Vision},
    booktitle = {Proceedings of the International Conference on Intelligent Robots and Systems (IROS)},

@article{peri2023empirical,

@misc{liu2024selfsupervisedmultiframeneuralscene,
  pages={13126--13136},
  booktitle={2019 IEEE/CVF International Conference on Computer Vision (ICCV)}, 
  title={{Exploiting Rigidity Constraints for LiDAR Scene Flow Estimation}},
and Brostow, Gabriel
year={2023}, 
  booktitle={Computer Vision--ECCV 2020: 16th European Conference, Glasgow, UK, August 23--28, 2020, Proceedings, Part II 16},
year = {2023},
  title={{Joint 3d estimation of vehicles and scene flow}},
  author={Radford, Alec and Wu, Jeff and Child, Rewon and Luan, David and Amodei, Dario and Sutskever, Ilya},
  year={2023}
  title={{Motion-based object segmentation based on dense rgb-d scene flow}},
    title         = "Velodyne Lidar Alpha Prime",
and Wang, Jie
  year={2009},
  volume={},
    journal   = {Twelfth International Conference on Learning Representations (ICLR)},
  author = {Radford, Alec and Narasimhan, Karthik and Salimans, Tim and Sutskever, Ilya},

@inproceedings{vedder2022sparse,

@article{chodosh2023,

@String(NIPS= {Adv. Neural Inform. Process. Syst.})

@String(WACV= {Wint. Conf. App. Comput. Vis})

@String(ICCV= {Int. Conf. Comput. Vis.})

@inproceedings{vedder2024zeroflow,
  title={Self-supervised learning of non-rigid residual flow and ego-motion},
  author={Mayer, Nikolaus and Ilg, Eddy and Hausser, Philip and Fischer, Philipp and Cremers, Daniel and Dosovitskiy, Alexey and Brox, Thomas},

@inproceedings{waymoopen,

@InProceedings{flowssl,
  pages={150--159},

@inproceedings{objectdetectionmotion,
  title={{Multi-view stereo reconstruction and scene flow estimation with a global image-based matching score}},
  pages={12265--12274},
  author    = {Kirmayr, Johannes and Wulff, Philipp},
              Dmitry Lagun and Issam Laradji and Hsueh-Ti (Derek) Liu and Henning Meyer and
  author={Quiroga, Julian and Brox, Thomas and Devernay, Fr{\'e}d{\'e}ric and Crowley, James},

@inproceedings{khurana2023point,
  author={Huguet, Fr{\'e}d{\'e}ric and Devernay, Fr{\'e}d{\'e}ric},
  booktitle=WACV,

@article{cbgs,

@misc{flow4d,
and {\c{C}}ayl{\i}, Y{\i}lmaz Kaan

@InProceedings{fastnsf,
author = {Rajeswaran, Aravind and Kumar, Vikash and Finn, Chelsea and Gupta, Abhinav},

@article{flownet3d,

@inproceedings{kittenplon2021flowstep3d,
  title        = {{The effectiveness of {MAE} pre-pretraining for billion-scale pretraining}},
  title={Tracking Without Bells and Whistles}, 
  journal=CVIU,
  title={{Smooth shells: Multi-scale shape registration with functional maps}},
    author        = {Arjun Majumdar and Karmesh Yadav and Sergio Arnaud and Yecheng Jason Ma and Claire Chen and

@String(IV = {Intel. Veh. Symp. IV})

@inproceedings{li2021hcrf,
  title={{Scene flow from point clouds with or without learning}},
                  Christoph Feichtenhofer and
      title={{Revisiting Weakly Supervised Pre-Training of Visual Perception Models}}, 
  year={2018},
  journal = {arXiv:2111.06377},
                  Priya Goyal and
  booktitle = {Proceedings of the Neural Information Processing Systems Track on Datasets and Benchmarks (NeurIPS Datasets and Benchmarks 2021)},
    year={2023}
title = {{Masked Autoencoding for Scalable and Generalizable Decision Making}},
 volume = {33},
 booktitle = {Advances in Neural Information Processing Systems},
title = {{3D Multi-Object Tracking: A Baseline and New Evaluation Metrics}}, 
  journal=NIPS,
location = {Portland, Oregon},
    website={http://vedder.io/zeroflow.html},
journal={ArXiv},
  pages={1765--1770},
  journal   = {CoRR},
  journal = {arXiv preprint arXiv: Arxiv-2305.16291}

@inproceedings{behl2019pointflownet,
  journal={arXiv preprint arXiv:2407.01702},
  title={{A variational method for scene flow estimation from stereo sequences}},

@article{lin2024icp,
  pages={145--155},

@String(CVIU = {Comput. Vis. Img. Und.})

@String(AAAI = {AAAI})

@article{peri2022towards,
  year    = {2022}
year = {2023}
  journal   = RAL,
  author={Jiang, Lu and Huang, Di and Liu, Mason and Yang, Weilong},
  title={{Implicit neural representations with periodic activation functions}},
  pages={12776--12785},
title = {{PointOdyssey: A Large-Scale Synthetic Dataset for Long-Term Point Tracking}},

@inproceedings{dewan2016rigid,
  title={{Adam: A method for stochastic optimization}},
  journal=PAMI,
  author={Chen, Zhiqin and Zhang, Hao},
  title={{Meteornet: Deep learning on dynamic 3d point cloud sequences}},
  title     = {{Robust Dynamic Radiance Fields}},
  year={2015}
  year    = {2023},
      title={Technical Report for Argoverse Challenges on Unified Sensor-based Detection, Tracking, and Forecasting},

@article{zhang2024deflow,

@inproceedings{puy2020flot,
  title={{Dynamic 3D Scene Analysis by Point Cloud Accumulation}},
  title={{SLIM: Self-supervised LiDAR scene flow and motion segmentation}},
              Yishu Miao and Derek Nowrouzezahrai and Cengiz Oztireli and Etienne Pot and

@inproceedings{argoverse2,
month = {07},
 title = {{ImageNet Classification with Deep Convolutional Neural Networks}},
 year = {2020}
  month     = {June},
  author    = {Huang, Xiangru and Wang, Yue and Guizilini, Vitor Campagnolo and Ambrus, Rares Andrei and Gaidon, Adrien and Solomon, Justin},


%% file: references.bib
@INPROCEEDINGS{ntp,
  author={Wang, Chaoyang and Li, Xueqian and Pontes, Jhony Kaesemodel and Lucey, Simon},
  booktitle={CVPR}, 
  title={{Neural Prior for Trajectory Estimation}}, 
  year={2022},
  volume={},
  number={},
  pages={6522-6532},
  doi={10.1109/CVPR52688.2022.00642}}

@inproceedings{khatri2024trackflow,
    author = {Khatri, Ishan and Vedder, Kyle and Peri, Neehar and Ramanan, Deva and Hays, James},
    title = {{I Can't Believe It's Not Scene Flow!}},
    booktitle = {European Conference on Computer Vision (ECCV)},
    year = {2024},
    website = {http://vedder.io/trackflow.html}
}

@article{peri2022futuredet,
  title={{Forecasting from LiDAR via Future Object Detection}},
  author={Peri, Neehar and Luiten, Jonathon and Li, Mengtian and Osep, Aljosa and Leal-Taixe, Laura and Ramanan, Deva},
  journal={arXiv:2203.16297},
  year={2022},
}

@article{surveyofpersonrecog,
author = {Nalty, Chrisopher and Peri, Neehar and Gleason, Joshua and Castillo, Carlos and Hu, Shuowen and Bourlai, Thirimachos and Chellappa, Rama},
year = {2022},
month = {12},
pages = {},
title = {{A Brief Survey on Person Recognition at a Distance}},
doi = {10.48550/arXiv.2212.08969}
}

@misc{
ramasinghe2024on,
title={{On the Optimality of Activations in Implicit Neural Representations}},
author={Sameera Ramasinghe and Hemanth Saratchandran and Violetta Shevchenko and Alexander Long and Simon Lucey},
year={2024},
url={https://openreview.net/forum?id=0Lqyut1y7M}
}

@article{nerf,
author = {Mildenhall, Ben and Srinivasan, Pratul P. and Tancik, Matthew and Barron, Jonathan T. and Ramamoorthi, Ravi and Ng, Ren},
title = {{NeRF: representing scenes as neural radiance fields for view synthesis}},
year = {2021},
issue_date = {January 2022},
publisher = {Association for Computing Machinery},
address = {New York, NY, USA},
volume = {65},
number = {1},
issn = {0001-0782},
journal = {Commun. ACM},
month = {dec},
pages = {99–106},
numpages = {8}
}

@inproceedings{garf,
author = {Chng, Shin-Fang and Ramasinghe, Sameera and Sherrah, Jamie and Lucey, Simon},
title = {{Gaussian Activated Neural Radiance Fields for High Fidelity Reconstruction and Pose Estimation}},
year = {2022},
isbn = {978-3-031-19826-7},
publisher = {Springer-Verlag},
address = {Berlin, Heidelberg},
booktitle = {Computer Vision – ECCV 2022: 17th European Conference, Tel Aviv, Israel, October 23–27, 2022, Proceedings, Part XXXIII},
pages = {264–280},
numpages = {17},
location = {Tel Aviv, Israel}
}

@inproceedings{vidanapathirana2023mbnsf,
  title={Multi-Body Neural Scene Flow},
  author={Vidanapathirana, Kavisha and Chng, Shin-Fang and Li, Xueqian and Lucey, Simon},
  booktitle={2024 International Conference on 3D Vision (3DV)},
  year={2024},
  organization={IEEE}
}

@inproceedings{neuralode,
author = {Chen, Ricky T. Q. and Rubanova, Yulia and Bettencourt, Jesse and Duvenaud, David},
title = {{Neural ordinary differential equations}},
year = {2018},
address = {Red Hook, NY, USA},
booktitle = {Proceedings of the 32nd International Conference on Neural Information Processing Systems},
pages = {6572–6583},
numpages = {12},
location = {Montr\'{e}al, Canada},
series = {NeurIPS'18}
}

@misc{chodosh2024simultaneousmapobjectreconstruction,
      title={{Simultaneous Map and Object Reconstruction}}, 
      author={Nathaniel Chodosh and Anish Madan and Deva Ramanan and Simon Lucey},
      year={2024},
      eprint={2406.13896},
      archivePrefix={arXiv},
      primaryClass={cs.CV},
      url={https://arxiv.org/abs/2406.13896}, 
}

@inproceedings{agro2024uno,
    title     = {{UnO: Unsupervised Occupancy Fields for Perception and Forecasting}},
    author    = {Agro, Ben and Sykora, Quin and Casas, Sergio and Gilles, Thomas and Urtasun, Raquel},
    booktitle = {CVPR},
    year      = {2024},
    }

@article{liquidneuralnet, title={{Liquid Time-constant Networks}}, volume={35},  journal={Proceedings of the AAAI Conference on Artificial Intelligence}, author={Hasani, Ramin and Lechner, Mathias and Amini, Alexander and Rus, Daniela and Grosu, Radu}, year={2021}, month={May}, pages={7657-7666} }

@inproceedings{Butler:ECCV:2012,
title = {A naturalistic open source movie for optical flow evaluation},
author = {Butler, D. J. and Wulff, J. and Stanley, G. B. and Black, M. J.},
booktitle = {European Conf. on Computer Vision (ECCV)},
editor = {{A. Fitzgibbon et al. (Eds.)}},
publisher = {Springer-Verlag},
series = {Part IV, LNCS 7577},
month = oct,
pages = {611--625},
year = {2012}
}

@inproceedings{
venkatesh2023samplingbased,
title={{Sampling-Based Model Predictive Control for Contact-Rich Manipulation}},
author={Sharanya Venkatesh and Bibit Bianchini and Alp Aydinoglu and Michael Posa},
booktitle={IROS 2023 Workshop on Leveraging Models for Contact-Rich Manipulation},
year={2023},
}

@inproceedings{weng2022fabricflownet,
  title={Fabricflownet: Bimanual cloth manipulation with a flow-based policy},
  author={Weng, Thomas and Bajracharya, Sujay Man and Wang, Yufei and Agrawal, Khush and Held, David},
  booktitle={Conference on Robot Learning},
  pages={192--202},
  year={2022},
  organization={PMLR}
}

@article{park2021nerfies,
  author    = {Park, Keunhong and Sinha, Utkarsh and Barron, Jonathan T. and Bouaziz, Sofien and Goldman, Dan B and Seitz, Steven M. and Martin-Brualla, Ricardo},
  title     = {{Nerfies: Deformable Neural Radiance Fields}},
  journal   = {ICCV},
  year      = {2021},
}

@INPROCEEDINGS{dynamicfusion,
  author={Newcombe, Richard A. and Fox, Dieter and Seitz, Steven M.},
  booktitle={2015 IEEE Conference on Computer Vision and Pattern Recognition (CVPR)}, 
  title={{DynamicFusion: Reconstruction and tracking of non-rigid scenes in real-time}}, 
  year={2015},
  volume={},
  number={},
  pages={343-352},
  doi={10.1109/CVPR.2015.7298631}}

@article{hu2024DepthCrafter,
            author      = {Hu, Wenbo and Gao, Xiangjun and Li, Xiaoyu and Zhao, Sijie and Cun, Xiaodong and Zhang, Yong and Quan, Long and Shan, Ying},
            title       = {{DepthCrafter: Generating Consistent Long Depth Sequences for Open-world Videos}},
            journal     = {arXiv preprint arXiv:2409.02095},
            year        = {2024}
    }
